\documentclass[journal]{IEEEtran}

\usepackage{cite}
\usepackage{graphicx}
\usepackage{amsmath, amssymb}
\usepackage[dvipsnames]{xcolor}
\usepackage{caption}
\usepackage{subcaption}
\usepackage[normalem]{ulem}
\usepackage[noend]{algorithm2e}
\SetAlCapSkip{1em}

\SetCommentSty{mycommfont}

\newtheorem{remark}{Remark}[section]
\newtheorem{theorem}{Theorem}[section]
\newtheorem{definition}[theorem]{Definition}
\newtheorem{problem}{Problem}[section]

\newcommand{\mymod}[1]{\textrm{#1}}
\newcommand {\myuline}{\bgroup \markoverwith
      {\color{MidnightBlue}{\rule[0.5ex]{2pt}{0.4pt}}}\ULon}
      
\DeclareMathOperator*{\argmin}{arg\,min}

\title{A Sequential Composition Framework for Coordinating Multi-Robot Behaviors}

\author{Pietro~Pierpaoli, Anqi~Li, Mohit~Srinivasan, Xiaoyi Cai, Samuel~Coogan, and Magnus~Egerstedt

\thanks{This work was supported by DARPA Grant No. N66001-17-2-4059.

Pietro Pierpaoli, Mohit Srinivasan, Samuel Coogan, and Magnus Egerstedt are with the School of Electrical and Computer Engineering, Georgia Institute of Technology, Atlanta, GA 30332, USA. (email: \{pietro.pierpaoli,mohit.srinivasan,sam.coogan,magnus\}@gatech.edu)

Anqi Li is with the department of Computer Science, University of Washington, Seattle, WA 98195, USA. (email:anqil4@cs.washington.edu)

Xiaoyi Cai is with the department of Aeronautics and Astronautics, Massachusetts Institute of Technology, Cambridge, MA 02139, USA. (email: xyc@mit.edu)

This paper has supplementary downloadable material available at http://ieeexplore.ieee.org, provided by the authors. This includes a multimedia MP4 format movie (60.1 MB), which shows commented implementation of the "Securing a Building" mission on a team of differential drive robots.
}
}

\begin{document}
\maketitle

\begin{abstract}
A number of coordinated behaviors have been proposed for achieving specific tasks for multi-robot systems. However, since most applications require more than one such behavior, one needs to be able to compose together sequences of behaviors while respecting local information flow constraints. Specifically, when the inter-agent communication depends on inter-robot distances, these constraints translate into particular configurations that must be reached in finite time in order for the system to be able to transition between the behaviors. To this end, we develop a \mymod{distributed} framework based on finite-time convergence control barrier functions that \mymod{enables a team of robots to adjust its configuration in order to meet the communication requirements for the different tasks.} In order to demonstrate the significance of the proposed framework, we implemented a \mymod{full-scale} scenario where a team of eight planar robots explore an urban environment in order to localize and rescue a subject.
\end{abstract}


\IEEEpeerreviewmaketitle

\section{Introduction} \label{sec:intro}
\IEEEPARstart{A}{s} our understanding of how to structure control and coordination protocols for teams of robots increases, a number of application domains have been identified, such as entertainment~\cite{ackerman2014flying}\cite{du2018fast}, surveillance~\cite{santos2018coverage}~\cite{shishika2018local}, manipulation~\cite{han2018hybrid}, and search-and-rescue~\cite{suarez2011survey}. Along with a decrease in the production and manufacturing costs associated with the platforms themselves, these applications have been enabled by a number of theoretical results that have emerged at the intersection of different disciplines such as robotics, controls, computer science, and graph theory~\cite{zelazo2018graph}. 

\begin{figure}
	\begin{center}
		\includegraphics[trim={2cm 2cm 1cm 2cm},width=0.35\columnwidth]{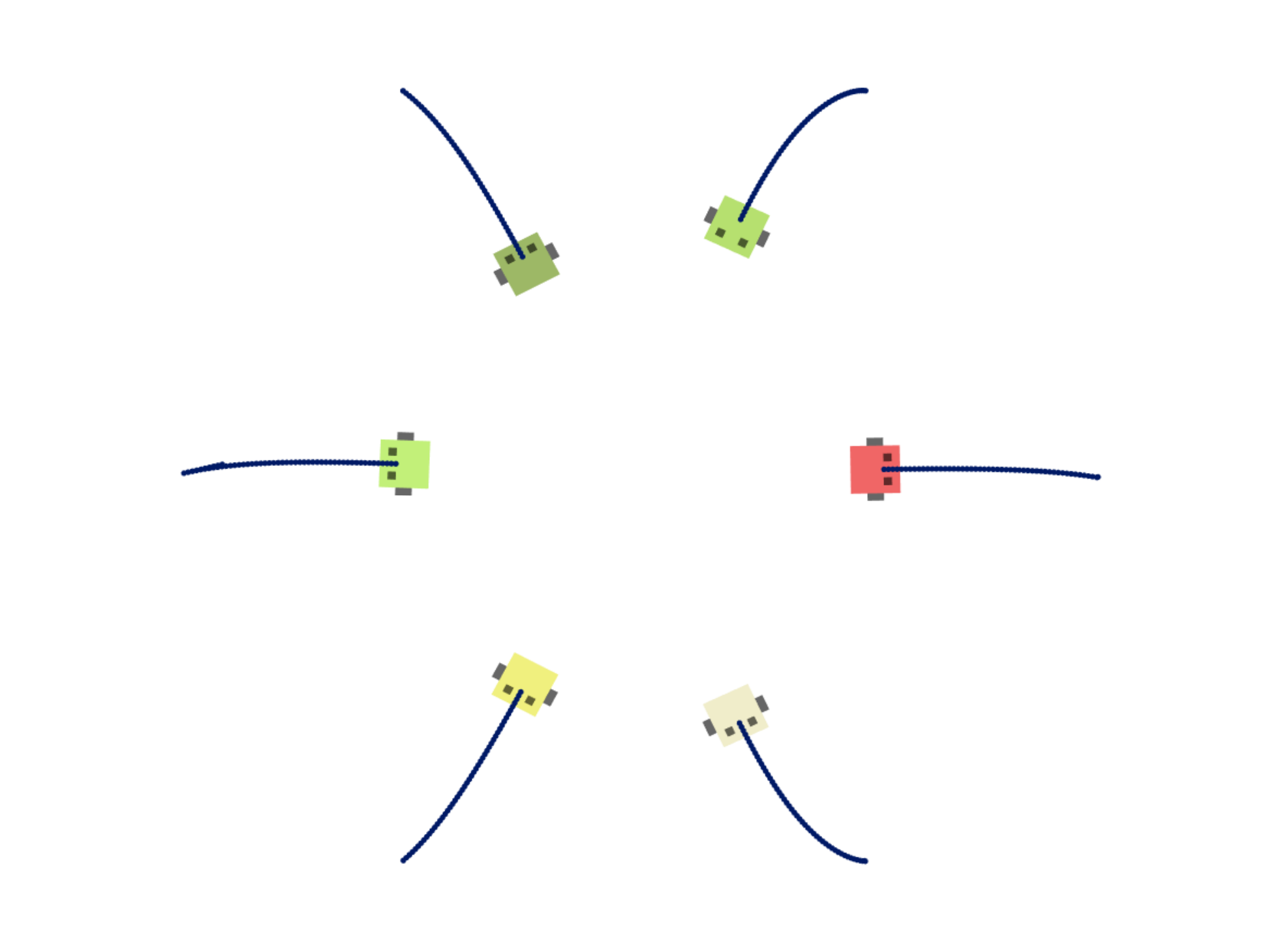}~
		\includegraphics[trim={3cm 2cm 2.5cm 2cm},width=0.32\columnwidth]{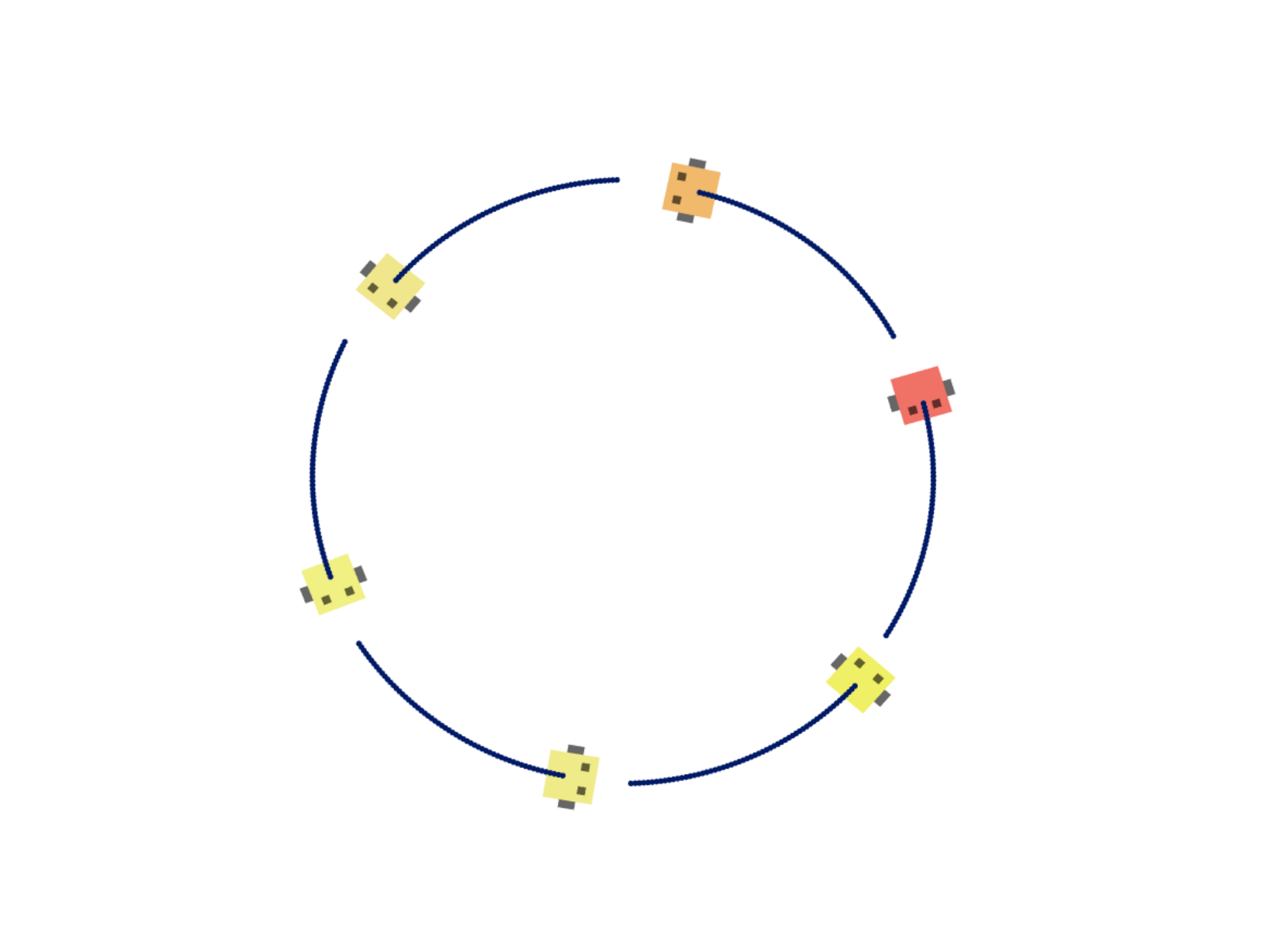}~
		\includegraphics[trim={2cm 2cm 2cm 2cm},width=0.32\columnwidth]{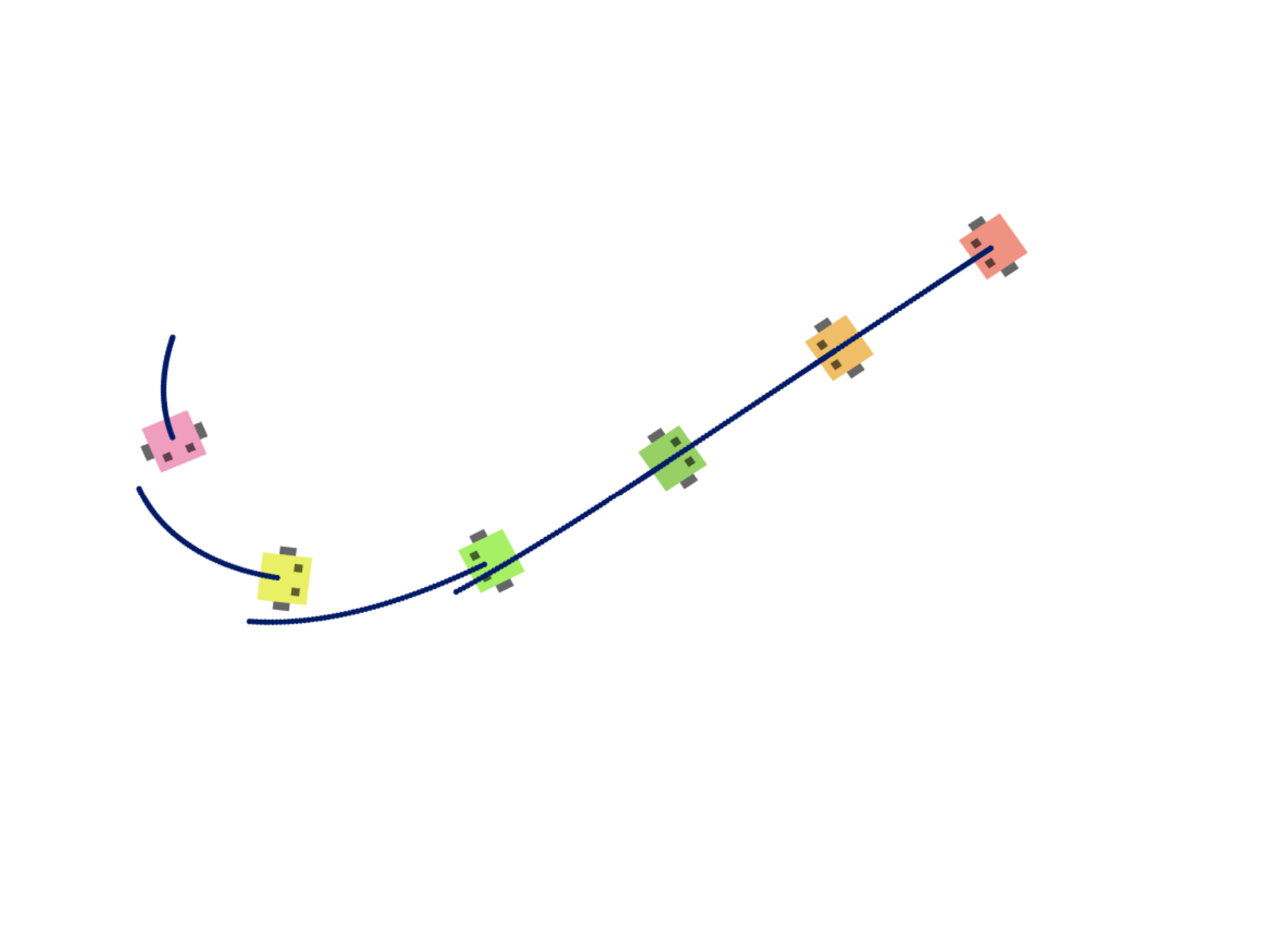}
		\caption{Simulation of three distributed multi-agent behaviors on a group of differential drive robots. From the left: rendezvous, cyclic-pursuit, and leader-follower. Solid lines indicate the past trajectories of the robots. \label{fig:coordBehExamples}}
	\end{center}
\end{figure}

From a motion controls perspective, one notable requirement is given by the need to define actions capable to solve team-wise objectives on the basis of locally available information. For instance, different extensions of the consensus equation have been used to arrive at locally defined controllers with provable, global convergence properties~\cite{cortes2017coordinated}. In this way, it is possible to construct coordinated controllers for the solution of motion control problems, such as rendezvous~\cite{lin2003multi}~\cite{ren2005coordination}, cyclic pursuit~\cite{ramirez2009cyclic}, formation control~\cite{lawton2003decentralized}~\cite{buckley2017infinitesimally}, coverage~\cite{cortes2004coverage}~\cite{santos2018coverage}, leader-based control~\cite{mesbahi2010graph}, and flocking~\cite{tanner2007flocking}. Particular instantiations of some of these behaviors are shown in Fig.~\ref{fig:coordBehExamples} on a group of six simulated differential drive robots.

For the correct execution of the controllers mentioned, a sufficiently rich set of information needs to be available to the robots. Representing the flow of information between the robots through {\it graphs}, with vertices and edges being respectively the robots and the pair-wise ability of sharing information, those conditions can be encoded in terms of particular graphs that need to exist between the robots. For example, rendezvous requires a spanning out-branching tree~\cite{mesbahi2010graph}, cyclic-pursuit requires a cyclic graph~\cite{ramirez2009cyclic}, formation control a rigid graph~\cite{mesbahi2010graph}, and a Delaunay graph is required for most of coverage control problems~\cite{cortes2004coverage}. 

Even though the coordinated behaviors mentioned above can address a number of different tasks, they have limited utility in the context of real-world missions, which can rarely be represented as single tasks. However, the utility of these behaviors can be greatly expanded if they are sequenced together, which is the primary consideration in this paper. But, for a construction like this to work, it is necessary that the required information is available to the robots as they transition from one behavior to the next.

As such, the problem of composing different behaviors, can be recast in terms of the ability of the robots to establish the interactions needed at each stage of a mission. In particular, when the communication between agents depends on their relative configurations (e.g. relative distance or orientation), realizing a certain communication structure directly affects the configuration of the system, which in turn, affects the execution of the mission itself. In order to overcome this coupling, we separate the problem of generating a sequence of behaviors that corresponds to the solution of a mission objective (e.g.~\cite{nagavalli2017automated}) from their composition. In this work we focus on the problem of designing a composition framework given a sequence of coordinated behaviors. \mymod{Although the focus of this paper is confined to motion control tasks, our framework is applicable to other forms of autonomous collaboration where desired interaction structures between the robots are required by the mission, e.g., sharing of resources in heterogeneous teams~\cite{ramachandran2019resilience} or coordinated manipulation~\cite{culbertson2018decentralized}}.

The contribution of this paper is twofold. Firstly, extending the results in~\cite{li2018formally}, we propose a fully decentralized framework for composing a given sequence of multi-robot coordinated behaviors. Secondly, responding to the lack of established large-scale scenarios for the testing of multi-robot techniques, we propose a scenario called {\it Securing a Building}, which is rich and complex enough to capture many challenges and objectives of real-world implementations. \mymod{The significance of our framework is demonstrated through implementation of the Securing a Building scenario on a team of mobile robots.}

The remaining of this paper is organized as follows. In Section~\ref{sec:fcbf} we review the definition of finite-time convergence barrier functions, while in Section~\ref{sec:problem} we present a centralized multi-robot composition framework, which is extended to a \mymod{fully decentralized} formulation in Section~\ref{sec:multAgImp}. In Section~\ref{sec:securing}, we describe the {\it Securing a Building} case study and its implementation. Finally, motivated by the lack of well-established scenarios for testing and comparing multi-agent robotics techniques, \mymod{in Appendix~\ref{sec:appendixA} we discuss supportive arguments for considering the Securing a Building as a multi-robot benchmark scenario.}
\section{Related Work} 
The problem of partitioning complex objectives into simpler tasks can be solved by sequentially composing {\it primitives}, e.g.,~\cite{cassandras2009introduction}, or by blending them simultaneously in a {\it hierarchical} fashion. An example of hierarchical composition for single robot motion control is navigation between points while avoiding obstacles, e.g.,~\cite{arkin1998behavior}. \mymod{In general, the problem of controlling a system by composing different modes of operation pertains to hybrid systems and {\it multi-modal} control domains~\cite{koutsoukos2000supervisory}.}

Because of the complexity emerging from the composition of distinct controllers, guarantees on the safety and correctness of the final results need to be established~\cite{kress2018synthesis}. Provable correct composition of control policies is investigated in the formal methods literature. Recently, compositional strategies inspired from formal methods have been used for the development of control strategies for multi-robot systems~\cite{srinivasan2018control,garg2019control,meyer2019hierarchical,chen2018verifiable}. In particular, the authors of~\cite{srinivasan2018control} use tools from linear temporal logic (LTL) for the specification of behaviors to be executed by the system. The solution is based on a sequence of constrained reachability problems, each consisting of a target set to be reached in finite time and a safety set within which the system must stay at all times. \mymod{A related approach is developed in~\cite{garg2019control}, where the problem of prescribed-time convergence to spatio-temporal specifications is formulated using control barrier functions.} The authors in~\cite{meyer2019hierarchical} discuss a hierarchical decomposition method for controller synthesis given LTL specifications.

In the context of controllers composition for multi-robot systems, in~\cite{belta2007symbolic} the authors use symbolic methods in order to generate high-level instructions from form of {\it human-like} language. The authors of~\cite{klavins2000formalism} introduce a framework for the composition of controllers in robotic systems using Petri Nets. In~\cite{marino2009behavioral}, behaviors from the Null-Space-Behaviors framework are combined in order to solve ad-hoc tasks, such as perimeter patrol. A supervisor, represented as a finite state automata, selects high-level behaviors by assembling low-level behaviors. In~\cite{nagavalli2017automated}, a revised version of the $A^*$ algorithm is used to generate an optimal path of behaviors, such that the overall cost of the mission is minimized. Similarly, in~\cite{vukosavljev2019hierarchically} motion planning for a team of quadcopters is solved by defining higher level motion primitives obtained by a spatial partition of the environment. However, none of these approaches specifically address the problem of correct composition between primitives, which is the focus of this paper.

As discussed in the previous section, coordination between agents is possible only if particular interactions exist between the robots. In multi-robot systems, interaction requirements are commonly investigated in terms of connectivity maintenance, i.e., a certain graph or node-connectivity needs to be guaranteed at all times. Methods employed in the solution to this problem include edge weight functions~\cite{ji2007distributed}, control rules based on estimate of algebraic connectivity~\cite{sabattini2013distributed}, hybrid control~\cite{zavlanos2009hybrid}, passivity~ \cite{igarashi2009passivity}, and barrier functions~\cite{wang2016multi}. If connectivity between agents needs to be guaranteed in non-nominal circumstances, resilient solutions must be in place as well, e.g.,
\cite{ramachandran2019resilience}, \cite{panerati2019robust}, and~\cite{varadharajan2019unbroken}. Notably, a technique based on graph process specifications for the sequential composition of different multi-agent controllers is discussed in~\cite{twu2010graph}. Similar to our work, the authors in~\cite{twu2010graph} bridge the gap between composition of controllers and the topology requirements by encoding requisites for each controller in terms of graphs.
However, while in \cite{twu2010graph} {\it incompatible} controllers are combined through the introduction of a {\it bridging} controller, in our approach controllers are minimally modified by the robots in order to satisfy upcoming requirements. Our approach significantly reduces the complexity of the composition process, \mymod{minimizes the energy spent by the robots to switch between behaviors}, and can accommodate additional constraints, such as inter-robot collisions and obstacles avoidance.

\section{Finite-Time Barrier Functions} \label{sec:fcbf}
In this section we review the general definition of Finite-time Convergence Control Barrier Function (FCBF) which was first introduced in~\cite{li2018formally} and inspired by the finite-time stability analysis for autonomous system introduced in~\cite{bhat2000finite}. Given a dynamical system operating in an open set $\mathcal{D} \subseteq \mathbb{R}^n$ and a set $\mathcal{C}\subset\mathcal{D}$, barrier functions~\cite{xu2015robustness} are Lyapunov-like functions that guarantee forward invariance of $\mathcal{C}$ with respect to the state of the system. In other words, if an appropriate barrier function exists, it can be used to show that if the state of a system is in $\mathcal{C}$ at some time, it will be in $\mathcal{C}$ thereafter. The concept of barrier functions was extended to Zeroing Control Barrier Functions (ZCBF) in~\cite{xu2015robustness}, where asymptotic convergence of the state to the set $\mathcal{C}$ was discussed. Thus, provided that an appropriate ZCBF exists, if the state of the system is not in $\mathcal{C}$ at some initial time, it will asymptotically converge to $\mathcal{C}$. 

As discussed in the introduction, before execution of a coordinated behavior, robots need to satisfy certain spatial configurations imposed by the behavior itself. Importantly asymptotic convergence to the correct configuration is not sufficient. In fact, if we consider $\mathcal{C}$ as the joint set of all initial configurations required for a particular behavior, the state must strictly belong to $\mathcal{C}$ for the behavior to work properly. Following this observation, the need for a finite-time convergence extension of the previous concepts becomes clear. In particular, denoting the state of the system as $x(t)\in\mathcal{D}$, we are interested in verifying the following conditions:
\begin{itemize}
	\item if $x(t_0)\in\mathcal{C}$, then $x(t)\in\mathcal{C}$ for all $t > t_0$
	\item if $x(t_0) \notin \mathcal{C}$, then $x(t)\in\mathcal{C}$ for some $t_0<t<\infty$.
\end{itemize}
In order to do this, we encode the set $\mathcal{C} \subset \mathcal{D} \subseteq \mathbb{R}^n$, through the superzero-level set of a continuously differentiable function $h: \mathcal{D} \rightarrow \mathbb{R}$, i.e.,
\begin{equation}
	\mathcal{C} = \{ x\in \mathcal{D} \, | \, h(x)\geq 0 \}.
	\label{eq:sefdef}
\end{equation}
\begin{definition}
We introduce the following class-$\mathcal{K}$ function
\begin{equation}
	\bar{\alpha}_{\rho,\gamma}(h(x)) = \gamma \cdot \,\text{sign}(h(x))\, \cdot |h(x)|^\rho,	\label{eq:alpha}
\end{equation}
with $\rho \in [0,1)$ and $\gamma>0$, which is continuous everywhere and locally Lipschitz everywhere except at the origin~\cite{bhat2000finite}.
\end{definition}

\begin{definition}~\cite{li2018formally} For a dynamical system 
\begin{equation} \label{eq:affinesystem}
	\dot{x} = f(x)+g(x)u
\end{equation} with $x\in\mathcal{D}$, $u \in U \subset \mathbb{R}^m$, and for a set $\mathcal{C}$ induced by $h$, if there exists a function $\bar{\alpha}_ {\rho,\gamma}(h(x))$ of the form~(\ref{eq:alpha}) such that
	\begin{equation}
		\sup_{u \in U}\bigg\{ L_fh(x) + L_gh(x)u + \bar{\alpha} _{\rho,\gamma}(h(x)) \bigg\} \geq 0 \quad \forall x\in\mathcal{D},
	\end{equation}
then, the function $h$ is a {\it Finite-time Convergence Barrier Function }(FCBF) defined on $\mathcal{D}$.
\end{definition}

Following from the definition above, we define the set of admissible control inputs as
\begin{equation} \label{eq:admisU}
	K(x) = \{ u \in U \, | \, L_fh(x) + L_gh(x)u + \bar{\alpha} _{\rho,\gamma}(h(x)) \geq 0\}.
\end{equation}

\begin{theorem} \label{thm:FCBF} \cite{li2018formally}~Given a set $\mathcal{C}\subset \mathbb{R}^n$, any Lipschitz continuous controller $\mathcal{U}: \mathcal{D} \mapsto U$ such that 
\begin{equation} \label{eq:fcbf_controllers}
	\mathcal{U}(x) \in K(x) \qquad \forall x\in\mathcal{D},
\end{equation}
renders $\mathcal{C}$ forward invariant for the system~(\ref{eq:affinesystem}). Moreover, given an initial state $x_0 \in \mathcal{D} \backslash \mathcal{C}$, the same controller $\mathcal{U}$ results in $x(T)\in\mathcal{C}$, where
	\begin{equation}
		T \leq \frac{1}{\gamma(1-\rho)}|h(x_0)|^{1-\rho}.
	\end{equation}
\end{theorem}

In conclusion, by selecting a controller that verifies condition~(\ref{eq:fcbf_controllers}), both forward invariance and finite-time convergence to the desired set are guaranteed. 


\section{Problem Formulation}
\label{sec:problem}
We denote the state of a team of $n$ homogeneous mobile robots operating in a $d$-dimensional and connected domain $\mathcal{D}$ as $x(t) = [x_1(t)^T,\dots,x_n (t)^T]^T \in \mathcal{D} \subset$ $\mathbb{R}^{dn}$ where $x_i(t)\in\mathbb{R}^d$ is the position of robot $i$ at time $t$. As part of the coordinated nature of the behaviors being performed by the robots, each robot executes a control protocol which depends on the state of the subset of robots with which it interacts. We assume robots can communicate if the distance between them is less or equal to a sensing threshold $\Delta\in\mathbb{R}_{>0}$. Thus, the list of possible interactions between agents are described by a time-varying, undirected, proximity graph $\mathcal{G}(t)=(V,E(t))$, where $V=\{1,\dots,n \}$ is the set of nodes representing the robots and $E(t)$ is the set of interacting pairs at time $t$, where 
\begin{equation}
E(t) = \{ (i,j) \in V \times V\, | \, \| x_i(t)-x_j(t)\| \leq \Delta \}.	
\end{equation}
For each robot $i=1,\dots,n$, we denote the set of available neighbors at time $t$ as $\mathcal{N}_i(t) = \{ j \in V\, | \, (i,j) \in E(t) \}$, which depends on the position of the robots at time $t$.

The ensemble dynamics of the multi-agent system is described by
\begin{equation}
	\dot{x} = f(x) + g(x)\,u
	\label{eq:ensembleDynamics}
\end{equation}
where $f$ and $g$ are continuous locally Lipschitz continuous functions and $u = [u_1^T,\dots,u_n^T]^T \in U \subset \mathbb{R}^m$ is the vector of inputs, which depends on the particular behavior being executed. At all times, the control input $u$ in~(\ref{eq:ensembleDynamics}) is given by a controller $\mathcal{U}$, which can be defined as a state feedback law $\mathcal{U}: \mathcal{D} \mapsto U$ or by a combination of both external parameters and state feedback law $\mathcal{U}: \mathcal{D} \times \Theta  \mapsto U$, where $\Theta$ is a space of parameters appropriate for the behavior. For instance, the controller corresponding to a {\it weighted consensus} belongs to the first case. On the other side, a leader-follower protocol where followers maintain prescribed inter-agent distances is described by a controller that depends on both state feedback (followers' control) and exogenous parameters (leader's goal) (see Section~\ref{sec:multAgImp} for examples).

We represent a {\it mission} by an ordered sequence of $M$ coordinated behaviors
\begin{equation}
	\pi = \{ \mathcal{B}_1,\dots, \mathcal{B}_M\}.
\end{equation}
The $k^{\text{th}}$ behavior in $\pi$ is defined by the pair
\begin{equation}\label{eq:tuple}
	\mathcal{B}_k = \{ \mathcal{U}_k,\, \mathcal{G}_k\},
\end{equation}
where $\mathcal{U}_k$ represents the coordinated controller described above and $\mathcal{G}_k$ is the interaction graph required by behavior $\mathcal{B}_k$ to function properly. We assume the list of behaviors $\pi$ to be fixed and available to all robots. \mymod{We will use the term {\it behavior} to refer to a generalized multi-robot controller in the form~(\ref{eq:tuple}) and to {\it task} as the objective of the controller.}

\begin{figure}[h!]
\includegraphics[width=\columnwidth]{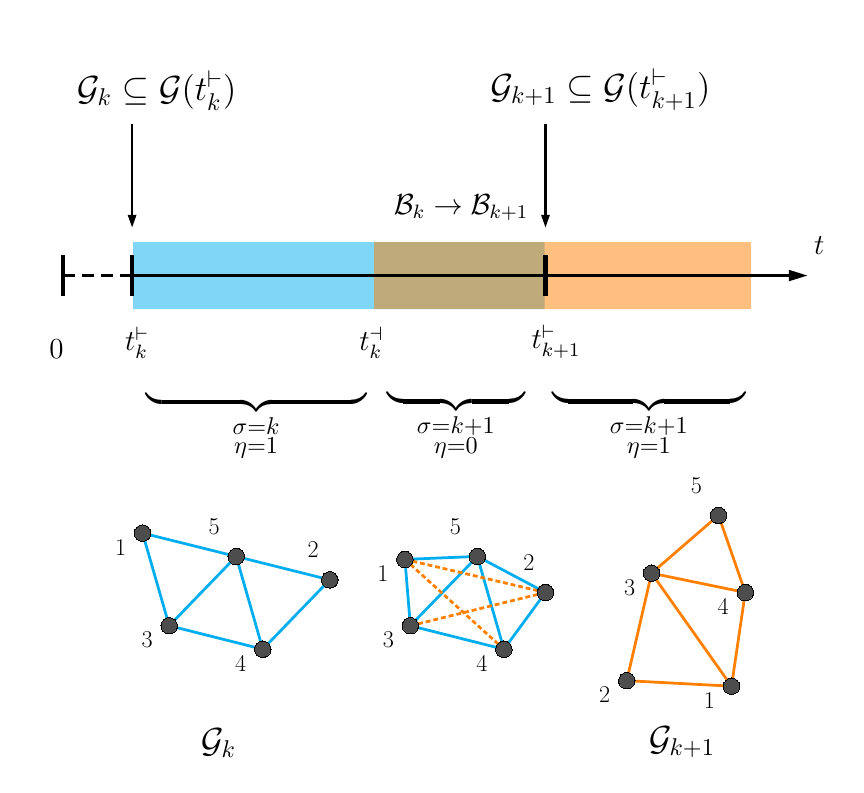}
  \caption{Schematic representation of the behaviors sequencing framework. Behavior $\mathcal{B}_k$ is executed during the blue portion of the timeline and $\mathcal{B}_{k+1}$ is executed during the orange portion. Sequential execution of behaviors requires each agent to reach a spatial configuration such that the desired graph is a spanning graph of the communication graph, i.e., $\mathcal{G}_k \subseteq \mathcal{G}(t_k^\vdash)$ and $\mathcal{G}_{k+1} \subseteq \mathcal{G}(t_{k+1}^\vdash)$ respectively. \label{fig:beh_seq}}
\end{figure}

As discussed in Section~\ref{sec:intro}, each behavior requires a certain interaction structure between the robots (i.e., pairs of robots that need to be neighbors). With reference to~(\ref{eq:tuple}), we describe an interaction structure via the graph $\mathcal{G}_k=(V,E_k)$. Thus, denoting by $t_k^\vdash$ and $t_k^\dashv$ the starting and ending times for behavior $k$, the robots' configuration needs to satisfy $\mathcal{G}_k \subseteq \mathcal{G}(t)$ for all $t \in [t_k^\vdash, t_k^\dashv]$. In other words, as shown in Fig.~\ref{fig:beh_seq}, the interaction structure required by each behavior needs to be a spanning graph of the graph induced by the state of the agents during the interval of time the behavior is executed. \mymod{At this point, given a list of behaviors constituting the mission $\pi$ and the corresponding multi-robot controllers, we want to design a procedure that enables robots to assemble and maintain the communication graph required by each behavior.}

\mymod{
\begin{problem} \label{pr:problem}
Given an ordered sequence of coordinated behaviors $\pi= \{ \mathcal{B}_1,\dots, \mathcal{B}_M\}$, where each $\mathcal{B}_k = \{ \mathcal{U}_k,\, \mathcal{G}_k\}$ can be completed by the robots in finite-time, design a feedback control policy to compose the behaviors such that
\begin{equation}
\mathcal{G}(t) \supseteq
    \begin{cases}
    \mathcal{G}_k \, &t \in [t_k^\vdash , t_k^\dashv] \\
    \mathcal{G}_k \cup \mathcal{G}_{k+1} \, &t \in (t_k^\dashv,t_{k+1}^\vdash)
\end{cases} \quad \forall \, k=1,\dots,M-1.
\end{equation}
\end{problem}
}
\section{Composition of Coordinated Behaviors}
\label{sec:seq_framework}

In addition to the list $\pi$, transitions between behaviors need to be synchronized, i.e., for each behavior $\mathcal{B}_k$, $k=1,\dots, M$, robots must 1) start assembling $\mathcal{G}_{k+1}$ only after all robots have completed $\mathcal{B}_k$ and 2) start executing $\mathcal{B}_{k+1}$ only after condition $\mathcal{G}_{k+1} \subseteq \mathcal{G}(t)$ is satisfied. We assume the existence of a discrete counter $\sigma \in [1,\dots,M]$ which indicates the active behavior and a binary signal
\begin{equation}
\eta(\sigma) = \begin{cases}
	1 \quad \text{if} \quad \mathcal{G}_k \subseteq \mathcal{G}(t) \\
	0 \quad \text{o.w.}
\end{cases}
\end{equation} 
which describes whether the interaction structure required by behavior $\mathcal{B}_\sigma$ is available. In this section, we assume both signals to be controlled by a supervisor and made available to the robots at all times, e.g., through a dedicated static communication network. \mymod{In the next section, we discuss the extension to a fully distributed framework.}

Following from the communication modality assumed for the robots, communication constraints can be expressed in terms of relative distance between the robots. In other words, behavior $\mathcal{B}_k$ can be correctly executed if, for all $t\in[t_k^\vdash,t_k^\dashv]$, all the distances between pairs in $E_k$ are below the proximity threshold $\Delta$. To this end, a convenient pair-wise connectivity FCBF can be defined as
\begin{equation}
	h_{ij}^c(x) = \Delta^2 - \| x_i - x_j \|^2,
	\label{eq:commBarriers}
\end{equation}
and we note that if $\|x_i-x_j\| \leq \Delta$, then $h_{ij}^c(x)\geq 0$.
In addition, the edge-level and ensemble-level connectivity constraint sets for behavior $\mathcal{B}_k$ are
\begin{align}
\mathcal{C}_{ij}^c &= \{ x \in \mathcal{D} \,|\, h_{ij}^c(x) \geq 0 \} \\
\mathcal{C}^c_k &= \{ x \in \mathcal{D} \,|\, h_{ij}^c(x) \geq 0, \, \forall (i,j)\in  E_k\}.
\end{align}

Following the definition given in~(\ref{eq:admisU}), the admissible set of control inputs that guarantees finite-time convergence to $\mathcal{C}^c_k$ is:	
\begin{multline} \label{eq:admGraphinput}
	K_k^c (x) = \{ u \in U \, | \, \dot{h}_{ij}^c(x) + \bar{\alpha} _{\rho,\gamma}(h_{ij}^c(x)) \geq 0 , \\ \forall (i,j)\in E_k \}
\end{multline}

\begin{theorem} \label{thm:fcbfControl}
	Denoting with $x_0$ the initial state of the system with dynamics~(\ref{eq:ensembleDynamics}), any controller $\mathcal{U}:\mathcal{D} \mapsto U$ such that $\mathcal{U}(x_0) \in K_k^c(x_0)$ for all $x_o \in \mathcal{D}$, will drive the system to $\mathcal{C}^c_k $ within time
\begin{equation} \label{eq:fcTime}
	T_k = \max_{ (i,j) \in E_k |  h^c_{ij}(x_0)<0} \left\{ \frac{1}{ \gamma(1-\rho)} | h_{ij}^c(x_0) |^{1-\rho} \right\}.\end{equation}
\end{theorem}

\begin{IEEEproof} 
	Consider all pairs of agents $i$ and $j$, such that $(i,j) \in E_k$. If $h_{ij}^c(x_0) \geq 0$, i.e., agents $i$ and $j$ are within communication distance, the forward invariance property of $\mathcal{U}$, guarantees that $i$ and $j$ will stay connected. In this case, the state will reach $\mathcal{C}_{ij}^c$, within time $T_{ij}=0$. On the other side, consider $h_{ij}^c(x_0)<0$. Any $\mathcal{U}(x_0) \in K_k^c(x_0)$ satisfies the finite-time convergence barrier certificates, and because of Theorem~\ref{thm:FCBF}, if $x_0 \notin \mathcal{C}_{ij}^c$, then $x(T_{ij})\in \mathcal{C}_{ij}^c $, with
\begin{equation}
T_{ij} \leq  \frac{1}{ \gamma(1-\rho)} | h_{ij}^c(x_0) |^{1-\rho}.
\end{equation}   
Since every communication constraint $\mathcal{C}_{ij}^c $ will be reached within time $T_{ij}$, the total time required to drive $x(t)$ to $\mathcal{C}_k^c$ is upper bounded by
\begin{equation}
	T_k = \max_{(i,j) \in E_k |  h^c_{ij}(x_0)<0} T_{ij}.
\end{equation}	
\end{IEEEproof} 
When selecting control inputs from set~(\ref{eq:admGraphinput}), the system~(\ref{eq:ensembleDynamics}) will satisfy requirements for behavior $\mathcal{B}_k$ in finite time.

\subsection{Finite-Time Convergence Control Barrier Functions}
\label{sec:ftcontrolBF}
Once behavior $\mathcal{B}_{k-1}$ is completed, robots are required to converge to the set $\mathcal{C}^c_{k}$ before behavior $\mathcal{B}_{k}$ can start. Under the lead of the external supervisor, the change of behavior is communicated to the robots through the signal $\sigma$, which transitions from $k-1$ to $k$ once $\mathcal{B}_{k-1}$ is completed. Now, although finite-time convergence to $\mathcal{C}^c_{k}$ can be achieved by selecting any control input in $K_{k}^c(x)$, we seek to minimally perturb the execution of the behavior just concluded, namely $\mathcal{B}_{k-1}$. This can be accomplished by solving a problem similar to the one proposed in~\cite{ames2014control}, which we adapt to our framework. Denoting with $\hat{u}_{k}=\mathcal{U}_{k}(x)$ the nominal control input from behavior $\mathcal{B}_{k}$, during transition between $\mathcal{B}_{k-1}$ and $\mathcal{B}_{k}$ the actual control input to the robots $u^*$ is defined as
\begin{equation}\label{eq:minQP1}
u^* = \argmin_{u \in U}  \| \hat{u}_{k-1} - u \|^2 \\
\end{equation}
subject to
\begin{equation} \label{eq:constrTransition}
   L_f\,h_{ij}^c + L_g\,h_{ij}^c\,u + \bar{\alpha}_{\rho,\gamma}(h_{ij}^c) \geq 0, 
\end{equation}
for all $(i,j) \in  E_{k-1} \cup E_{k}$. Once all required edges $E_{k}$ are established (i.e., $\eta=1$), edges in $E_{k-1}$ are no longer necessary. At this point, under the effect of the controller $\mathcal{U}_{k}$, the list of constraints in~(\ref{eq:constrTransition}) is substituted with
\begin{equation} \label{eq:constrExecution}
   L_f\,h_{ij}^c + L_g\,h_{ij}^c\,u + \bar{\alpha}_{\rho,\gamma}(h_{ij}^c) \geq 0,
\end{equation}
for all $(i,j)\in  E_{k}$. Since the cost function is convex and the inequality constraints~(\ref{eq:constrTransition}) and~(\ref{eq:constrExecution}) are control affine, the problem can be solved in real-time. In conclusion, because of the finite-time convergence and forward invariance properties of the above formulation, \mymod{if $\mathcal{B}_{k-1}$ can be completed and a solution to~(\ref{eq:minQP1}-\ref{eq:constrTransition}) (or (\ref{eq:minQP1}-\ref{eq:constrExecution})) exists}, robots will converge to the configuration required by $\mathcal{B}_{k}$, and maintain it throughout its execution.
\mymod{
\begin{remark}
The solution of~(\ref{eq:minQP1}-\ref{eq:constrTransition}) (or (\ref{eq:minQP1}-\ref{eq:constrExecution})) is contingent upon the existence of a control input capable to solve all constraints. In other words, $K_k^c(x) \cap K_{k+1}^c(x)$ (or $K_k^c(x)$) should not be empty for all times. For this, it is necessary that a robot's configuration that satisfies all constraints of the problem exists. However, this is not sufficient as the progress towards the desired configuration might be obstructed by constraints on the actuators or deadlock configurations. Although we do not address this directly, it is possible to mitigate feasibility issues by considering, for example, constraints relaxation, sum of squares barrier functions, or pre-defined back-up controllers (see~\cite{ames2019control} and references therein). \label{rmk:feasibility}
\end{remark}}

\subsection{Initial Constraints}
In addition to the communication constraints considered above, certain missions might require additional conditions to be met before each behavior can start. For example, during the exploration tasks it might be desirable for one robot to always stay within range of communication with a human-operator, or to maintain a minimum distance from an unsafe area. Assuming $\mathcal{B}_k$ requires a number of distinct $s_k$ of such constraints, we encode the entire set of initial conditions through a list of barrier functions $h_\ell^s(x)$, with $\ell = 1,\dots,s_k $:
\begin{equation} \label{eq:initialSet}
\mathcal{C}^{s}_k = \{ x \in \mathcal{D} \,|\, h_\ell^s(x) \geq 0, \, \forall \ell = 1,\dots,s_k \}.
\end{equation}
Following this definition, we define a set of admissible control inputs similar to the one in~(\ref{eq:admGraphinput}) that will drive the state of the system to the desired set within finite time:
\begin{multline} \label{eq:admInitCond}
	K_k^s (x) = \{ u\in U \, | \, \dot{h}_\ell^s(x) + \bar{\alpha} _{\rho,\gamma}(h_\ell^s(x)) \geq 0, \\
	 \forall \ell = 1,\dots,s_k \}.
\end{multline}

The set of controls satisfying both communication and initial conditions constraints can thus be obtained by intersection of set~(\ref{eq:admInitCond}) and~(\ref{eq:admGraphinput}):
\begin{equation}
	K_k(x) = K_k^c (x) \bigcap K_k^s (x).
\end{equation}

We note that the results in Theorem~\ref{thm:fcbfControl} and the formulation of minimally invasive controller in~(\ref{eq:minQP1}) still holds valid by considering the set $K_k(x)$ instead of $K_k^c (x)$ as the set of admissible control inputs.
\section{\mymod{Distributed Composition of Behaviors}}\label{sec:multAgImp}
The composition framework discussed in the previous section reduces to the team-wise minimum norm controller~(\ref{eq:minQP1}), which is not directly solvable by individual robots. \mymod{In addition to this, a centralized supervisor is needed in order to synchronize behavior transitions.} In this section, we \mymod{formulate a decentralized solution to problem~\ref{pr:problem} which can be implemented by the robots using only information from their neighbors. Furthermore, we also include those} additional constraints necessary for the safe operations of the robots, e.g., inter-agent collisions and obstacles avoidance~\cite{wang2016multi}. The formulation is derived following the approach described in~\cite{squires2019composition}, which we adapt here to our framework.

\subsection{Distributed Finite-Time Convergence Control Barrier Functions}
The limitation in solving problem~(\ref{eq:minQP1}) in a distributed fashion is represented by the fact that knowledge of dynamics, input $\hat{u}$, and state $x$ for the entire team need to be available. In addition, solution of~(\ref{eq:minQP1}), provides the control inputs for the entire team, which are unnecessary to the individual robots.

In order to develop the correct decentralized formulation of~(\ref{eq:minQP1}), we first define a decomposition of the dynamics~(\ref{eq:ensembleDynamics}). We denote by $\mathcal{D}_i\subset$ \mymod{$\mathbb{R}^d$} and $U_i \subset$ \mymod{$\mathbb{R}^m$} configuration space and set of feasible controls for agent $i$ respectively. In addition, by denoting with $\bar{f},\,\bar{g}: \mathcal{D}_i \mapsto$ \mymod{$\mathbb{R}^d$} the node-level terms of the control affine dynamics of agent $i$, the ensemble dynamics can be written as:
\begin{equation} \label{eq:decoup_dyn}
	\dot{x} = \bar{f}(x_i)\otimes {\bf 1}_n + (\bar{g}(x_i) \otimes I_n)\,\begin{bmatrix} u_1 \\ \vdots \\ u_n \end{bmatrix},
\end{equation}
where $u_i \in U_i$ is the $i^{\text{th}}$ robot's control input, $\otimes$ is  the Kronecker product, and ${\bf 1}_n$ and $I_n$ are vector of ones and identity matrix of size $n$ respectively.

Let's consider two sequential behaviors $\mathcal{B}_{k-1}$ and $\mathcal{B}_{k}$. Upon completion of $\mathcal{B}_{k-1}$, for all edges $(i,j)\in E_k$, robots' configuration should satisfy 
\begin{equation} \label{eq:dec_const_edge}
	\dot{h}_{ij}^c(x_i,x_j) + \bar{\alpha}_{\rho,\gamma}(h_{ij}^c(x_i,x_j)) \geq 0.
\end{equation}
From the $i^{\text{th}}$ robot's point of view, the set of constraints that need to be satisfied in order to execute the new behavior are
\begin{equation} \label{eq:dec_const_agenti}
	\dot{h}^c_{ij}(x_i,x_j) + \bar{\alpha}_{\rho,\gamma}(h^c_{ij}(x_i,x_j)) \geq 0 \quad \forall j\in \mathcal{N}_{k}^i,
\end{equation}
where we recall that $\mathcal{N}_{k}^i$ is the set of neighbors to robot $i$ required by behavior $\mathcal{B}_k$. However, since constraint~(\ref{eq:dec_const_agenti}) appears exactly twice across the team of robots, it can be relaxed by considering the admissible set of control inputs
\begin{equation} \label{eq:admGraphControl}
K_{k}^{c,i} = \bigcap_{j\in\mathcal{N}_k^i} K_{k,ij}^{c,i}
\end{equation}
with
\begin{equation}
K_{k,ij}^{c,i} = \{u_i\in U_i \,| \,L_{\bar{f}}h_{ij}^c + L_{\bar{g}}h_{ij}^cu_i + \frac{\bar{\alpha} _{\rho,\gamma}(h_{ij}^c)}{2} \geq 0 \},
\end{equation}
where dependence from $x_i$ and $x_j$ is omitted for clarity.

\begin{theorem} \label{thm:fcbfControl_dist}
	Denoting with $x_0 = [x_{0,1}^T,\dots,x_{0,n}^T]^T$ the initial state of a multi-agent system with dynamics described as in~(\ref{eq:decoup_dyn}), any controller $\mathcal{U}_i:\mathcal{D}_i^{|\mathcal{N}_k^i|} \mapsto U_i$ such that $\mathcal{U}_i(x_{0}) \in K_{k}^{c,i}$ for all $x_{0} \in \mathcal{D}_i^{|\mathcal{N}_k^i|}$, will drive the ensemble state to $\mathcal{C}^c_k $ within time
\begin{equation}
	T_k = \max_{\substack{(i,j) \in E_k \\ \text{s.t.} \,\, h^c_{ij}(x_{0,i},x_{0,j})<0}} \left\{ \frac{1}{ \gamma(1-\rho)} | h_{ij}^c(x_{0,i},x_{0,j}) |^{1-\rho} \right\}.
\end{equation}
\end{theorem}

\begin{IEEEproof}
From Theorem~\ref{thm:FCBF}, agents $i$ and $j$, with $(i,j)\in E_k$, will satisfy $h_{ij}^c \geq 0$ in finite time if
\begin{equation} \label{eq:connect_const}
\dot{h}_{ij}^c + \bar{\alpha}_{\rho,\gamma}(h_{ij}^c) \geq 0. 
\end{equation}

Considering the node level dynamics in~(\ref{eq:decoup_dyn}), the constraint~(\ref{eq:connect_const}) reduces to

\begin{equation} \label{eq:const2agents}
\begin{aligned}
&\frac{\partial h_{ij}^c}{\partial x_i} \left( \bar{f} + \bar{g}u_i \right) \, + \, \frac{\partial h_{ij}^c}{\partial x_j}\left(\bar{f} + \bar{g}u_j\right) + \bar{\alpha}_{\rho,\gamma}(h_{ij}^c) \geq 0 \\
&2\,L_{\bar{f}}h_{ij}^c + L_{\bar{g}}h_{ij}^c\,u_i + L_{\bar{g}}h_{ij}^c u_j + \bar{\alpha}_{\rho,\gamma}(h_{ij}^c) \geq 0
\end{aligned}
\end{equation}
which will be satisfied if both agents $i$ and $j$ satisfy the constraint 
\begin{equation} \label{eq:dec_const}
	\dot{h}_{ij}(x_i,x_j) + \frac{\bar{\alpha}_{\rho,\gamma}(h_{ij}(x_i,x_j))}{2} \geq 0.
\end{equation}
In addition, as discussed in Theorem~\ref{thm:fcbfControl},  constraint~(\ref{eq:const2agents}) will still be satisfied at time

\begin{equation}
	T_{ij} \leq \frac{1}{\gamma(1-\rho)} | h_{ij}^c(x_{0,i},x_{0,j}) |^{1-\rho}.
\end{equation}  

The same argument can be repeated for all pairs $(i,j) \in E_k$, and condition $\mathcal{G}_k \subseteq \mathcal{G}(t)$ will be satisfied within time

\begin{equation}
	T_k = \max_{\substack{(i,j) \in E_k \\ \text{s.t.} \,\, h^c_{ij}(x_{0,i},x_{0,j})<0}} \left\{ T_{ij} \right\}.
\end{equation}	
\end{IEEEproof}

Applying the same design principle described in Section~\ref{sec:ftcontrolBF}, the minimally invasive control action can be computed by each robot as
\begin{equation}\label{eq:minQP1_dist}
u^*_i = \argmin_{u_i \in U_i}  \| \hat{u}_{k-1,i} - u_i \|^2 \\
\end{equation}
subject to
\begin{equation}\label{eq:constrTransition_dist}
L_{\bar{f}}\,h_{ij}^c + L_{\bar{g}}\,h_{ij}^c\,u_i + \frac{\bar{\alpha}_{\rho,\gamma}(h_{ij}^c)}{2} \geq 0 , \quad \forall j\in  \mathcal{N}_{k-1}^i \cup \mathcal{N}_{k}^i.
\end{equation}
Similarly to constraint~(\ref{eq:constrTransition}), once all edges in $E_k$ are available, constraint~(\ref{eq:constrTransition_dist}) is substituted with
\begin{equation}\label{eq:constrExecution_dist}
L_{\bar{f}}\,h_{ij}^c + L_{\bar{g}}\,h_{ij}^c\,u_i + \frac{\bar{\alpha}_{\rho,\gamma}(h_{ij}^c)}{2} \geq 0 , \quad \forall j\in\mathcal{N}_{k}^i,
\end{equation}
which remains active until $\mathcal{B}_k$ is completed.

We note that, in order for agent $i$ to respect~(\ref{eq:constrExecution_dist}), the only external information needed is the state of all current neighbors, i.e. $x_j$ for all $j\in\mathcal{N}^i_k$. On the other side, in order to respect (\ref{eq:constrTransition_dist}), robots need to have access to the state of the future neighbors. This requirement can be satisfied through a state estimation scheme (e.g. EKF~\cite{williams2015observability}), which in turn requires knowledge of robots' dynamics (known for homogeneous teams) or network localization techniques~\cite{aspnes2006theory}. 
\mymod{
\begin{remark}
The ability of each robot to have access to an estimate of their future neighbors' state does not eliminate the necessity of establishing neighborhood relationships. In fact, a certain proximity structure between robots might be required by desired controllers' performance that cannot be met through state estimations, or by collaboration tasks that require physical interaction between the robots, e.g. collaborative manipulation~\cite{culbertson2018decentralized}, sharing of resources~\cite{ramachandran2019resilience}.
\end{remark}}

\subsection{Additional Constraints}
In addition to the proximity constraints discussed above, additional limitations might be imposed on the robots' configuration by the mission and the environment. For illustrative purposes, we consider inter-robots collisions and obstacle avoidance. Following the approach described in~\cite{wang2016multi}, we encode each pair-wise separation condition through the following barrier certificate
\begin{equation}
	h_{ij}^a(x) = \| x_i-x_j \|^2 - D_a^2 
\end{equation}
and the minimum separation $D_a$ between the robots is satisfied if $h_{ij}^a(x) \geq 0$, for all physical neighbors $j\in \mathcal{N}^i(t)$.

Similarly, avoidance of fixed obstacles can be introduce by considering $M$ ellipsoidal regions of the domain, described by centers $o=[o_1^T,\dots,o_M^T]^T$. For every agent-obstacle pair $(i,m)$ we define a pairwise barrier function as
\begin{align}
h_{im}^o(x) &=(x_i-o_m)^T\,P_m\,(x_i-o_m) - 1 \\
P_m &= \begin{bmatrix}
	a_m & 0 \\ 0 & b_m 
\end{bmatrix}
\quad a_m,b_m > 0.
\end{align}
The object avoidance constraints are satisfied if $h_{im}^o(x)\geq 0$, for all $i\in V$ and $m \in \{1,\dots,M \}=\mathcal{I}_M$.

Collecting all the constraints, we expand the problem formulation in~(\ref{eq:minQP1}) to

\begin{equation} \label{eq:minQP2}
\begin{aligned}
&\qquad u_i^* = \arg \min_{u_i \in U_i}  \| \hat{u}_{k-1,i} - u_i \|^2 &\\
&L_f\,h_{ij}^c + L_g\,h_{ij}^c\,u_i + \frac{\bar{\alpha}_{\rho,\gamma}(h_{ij}^c)}{2} \geq 0, & \forall j\in \mathcal{N}_{k}^i \\
&L_f\,h_{ij}^s + L_g\,h_{ij}^s\,u_i + \alpha(h_{ij}^{s}) \geq 0, &\forall j\in \mathcal{N}^i(t)\\
&L_f\,h_{im}^o + L_g\,h_{im}^o\,u_i + \alpha(h_{ij}^{s}) \geq 0, &\forall m \in \mathcal{I}_M
\end{aligned}
\end{equation}
where $\alpha$ is a locally Lipschitz extended class-$\mathcal{K}$ function and the first constraint is replaced by~(\ref{eq:constrTransition_dist}) during transitions. In conclusion, \mymod{if there exists a set of control inputs $u = [u_1,\dots,u_N]$ that simultaneously satisfies all constraints in~(\ref{eq:minQP2}), for all behaviors $k=1,\dots,M$, Problem~\ref{pr:problem} will be solved by the robots.}

\subsection{\mymod{Decentralized Behaviors Sequencing}}
\mymod{For the correct execution of the behaviors sequencing, each robot should start assembling a new graph only after all other robots have completed the current behavior. Similarly, a new behavior should start once all robots satisfy the neighbors' requirements for it. Now, we describe a decentralized strategy that allows execution of these two transitions without the need of a supervisor, nor synchronization between the robots.}

\begin{figure}
    \centering
    \includegraphics[width=0.8\columnwidth]{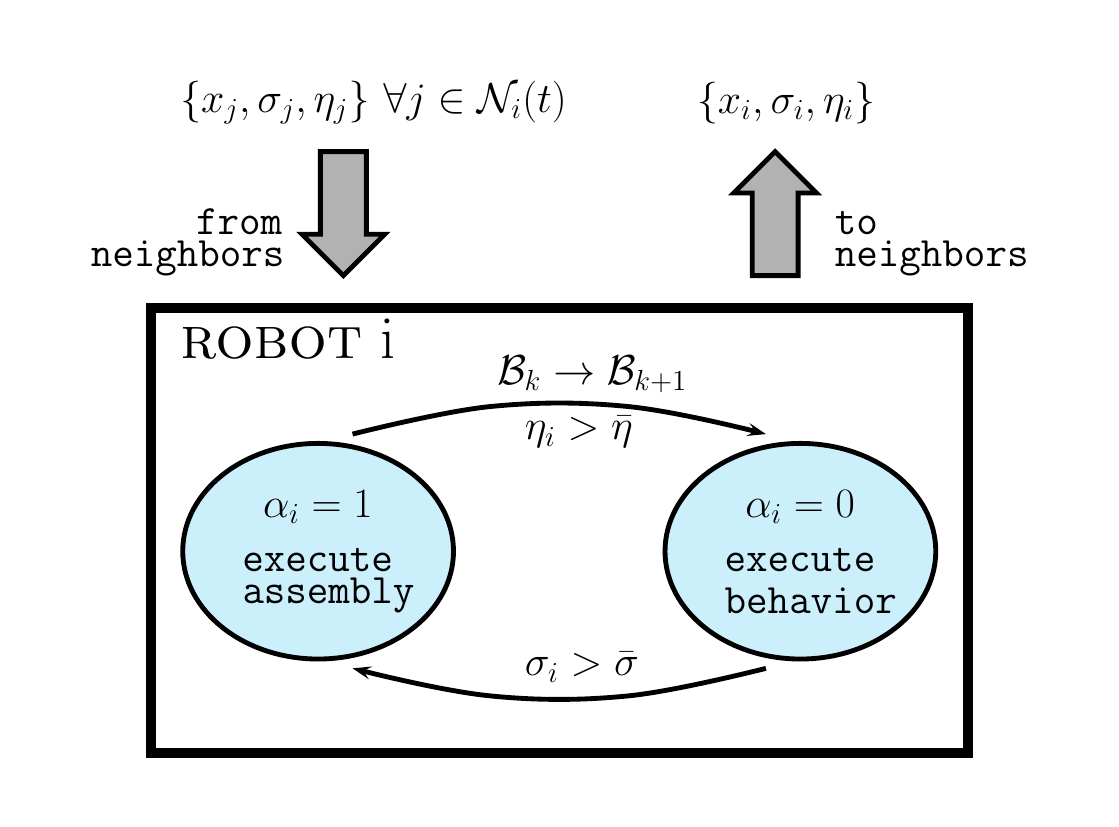}
    \caption{Representation of the distributed sequencing framework and information flow. At all times, each robot's state is in either {\tt behavior execution} ($\alpha_i=0$) or {\tt graph assembly} ($\alpha_i=0$) modes. Switching between the two modes is triggered by the variables $\sigma_i$ and $\eta_i$ whose values is continuously) updated through~(\ref{eq:sync}). When a switching between {\tt graph assembly} and {\tt behavior execution} occurs, a new behavior is started.}
    \label{fig:automata}
\end{figure}

\mymod{With reference to Fig.~\ref{fig:automata}, at any given time, each robot's mode of operation is described by a binary variable $\alpha_i$ that describes whether robot $i$ is assembling the graph for an upcoming behavior ($\alpha_i=1$) or executing a behavior ($\alpha_i=0$). Without loss of generality, assume robots' initial configuration satisfies the communication requirements for the first behavior, which is then executed ($\alpha_i = 0$). Once all robots have completed the first behavior, they start assembling the graph required by the following one ($\alpha_i = 1$), while minimally perturbing the behavior just concluded. Once the new graph is satisfied $\mathcal{G}_2 \subseteq \mathcal{G}(t)$, robots start behavior $\mathcal{B}_2$ and exit from assembly mode ($\alpha_i = 0$). This process repeats, until no successive behavior exists.} 

\mymod{A correct execution of this process requires robot to agree on when to perform transitions $\alpha_i = 0 \rightarrow 1$ and $\alpha_i = 1 \rightarrow 0$. To this end, we take inspiration from the consensus-based algorithm described in~\cite{wagenpfeil2009distributed} and we note that this choice is not central to the contribution of this paper. For each robot, we define a binary variable available only to robot $i$, $s_{t,i} \in \{0,1\}$ that denotes whether robot $i$ itself has completed its current task $s_{t,i} = 1$ ($s_{t,i} = 0$ if robot has not completed its current task). In addition, we introduce a variable $\sigma_i \in \mathbb{R}_+$, shared among neighbors, continuously updated through the following consensus-based process
\begin{equation} \label{eq:sync}
    \sigma_i^+ = s_{t,i} \frac{1}{|\mathcal{N}_i(t)|+1}\left(
    \sum_{j \in \mathcal{N}_i(t)} \sigma_j + 1 \right),
\end{equation}
where $\sigma_i^+$ represent the variable's value after the update. Owing to the diffusion of $\sigma_1,\dots,\sigma_N$ throughout the network, we can interpret $\sigma_i$'s as local measures of the team-wise completion of a task. As proved in~\cite{wagenpfeil2009distributed}, if $s_{t,i}=1$ for all $i=1,\dots N$ (i.e., all robots are capable to complete the current behavior), by following~(\ref{eq:sync}), $\lim_{t \rightarrow \infty} \sigma_i = 1$, for all $i=1,\dots N$. Therefore, robot $i$ starts assembling a new communication graph once the value of $\sigma_i$ is close enough to $1$ (see~\cite{wagenpfeil2009distributed} for a discussion on how to choose the switching threshold). A similar process is used for the transition $\alpha_i=1 \rightarrow 0$, where we replace $s_{t,i}$ and $\sigma_i$ with $s_{a,i}$ and $\eta_i$ respectively. The distributed sequencing procedure is summarized in Algorithm~\ref{alg:dis_sequence}.}

\begin{algorithm}[h]
\caption{Distributed composition of behaviors.} \label{alg:dis_sequence}
$\pi \gets \{\mathcal{B}_1,\dots, \mathcal{B}_M\}$ \tcc*[r]{initialize behaviors}
$k = 1$\;
$\alpha_i \gets 0$\;
\While{$k < M+1$}{
\tcc*[h]{Aggregate data from neighbors} \\
\For{$j \in \mathcal{N}_i(t)$}
{
$\{X_i, \Sigma_i, H_i\} \gets \{X_i, \Sigma_i, H_i\} \cup \{ x_j , \sigma_j , \eta_j \}$ \;
}
\tcc*[h]{Compute nominal control} \\
$\hat{u}_i \gets \mathcal{U}_k(x_i,X_i)$\; 
\tcc*[h]{Compute team-wise completion states} \\

\eIf{$\alpha_i == 0$}{
\lIf{{\tt task complete}}
{$s_e \gets 1$}
\lElse{$s_e \gets 0$} 
$\sigma_i:=s_e\,\frac{1}{|\mathcal{N}_k^i|+1}(\sum_{j \in \mathcal{N}_i(t)} \sigma_j + 1)$\;
\If{$\sigma_i > \bar{\sigma}$}{
$\alpha_i\gets 1$\; $k \gets k+1$ \;
}
}{ 
\lIf{{\tt assembly complete}}
{$s_a \gets 1$}
\lElse{$s_a \gets 0$} 
$\eta:=s_a\,\frac{1}{|\mathcal{N}_k^i|+1}(\sum_{j \in \mathcal{N}_i(t)} \eta_j + 1)$\;
\If{$\eta>\bar{\eta}$}{
$\alpha_i\gets 0$\; $s_e\gets 0$ \;}
}
\tcc*[h]{Solve FCBF QP} \\
$u_i \gets QP(\hat{u}_i,X_i,x_i)$
}
\end{algorithm}

\subsection{Applications}
\begin{figure*}[t]
\centerline{ 
\subcaptionbox{\label{fig2:a}}{\includegraphics[width=0.66\columnwidth, height=0.39\columnwidth]{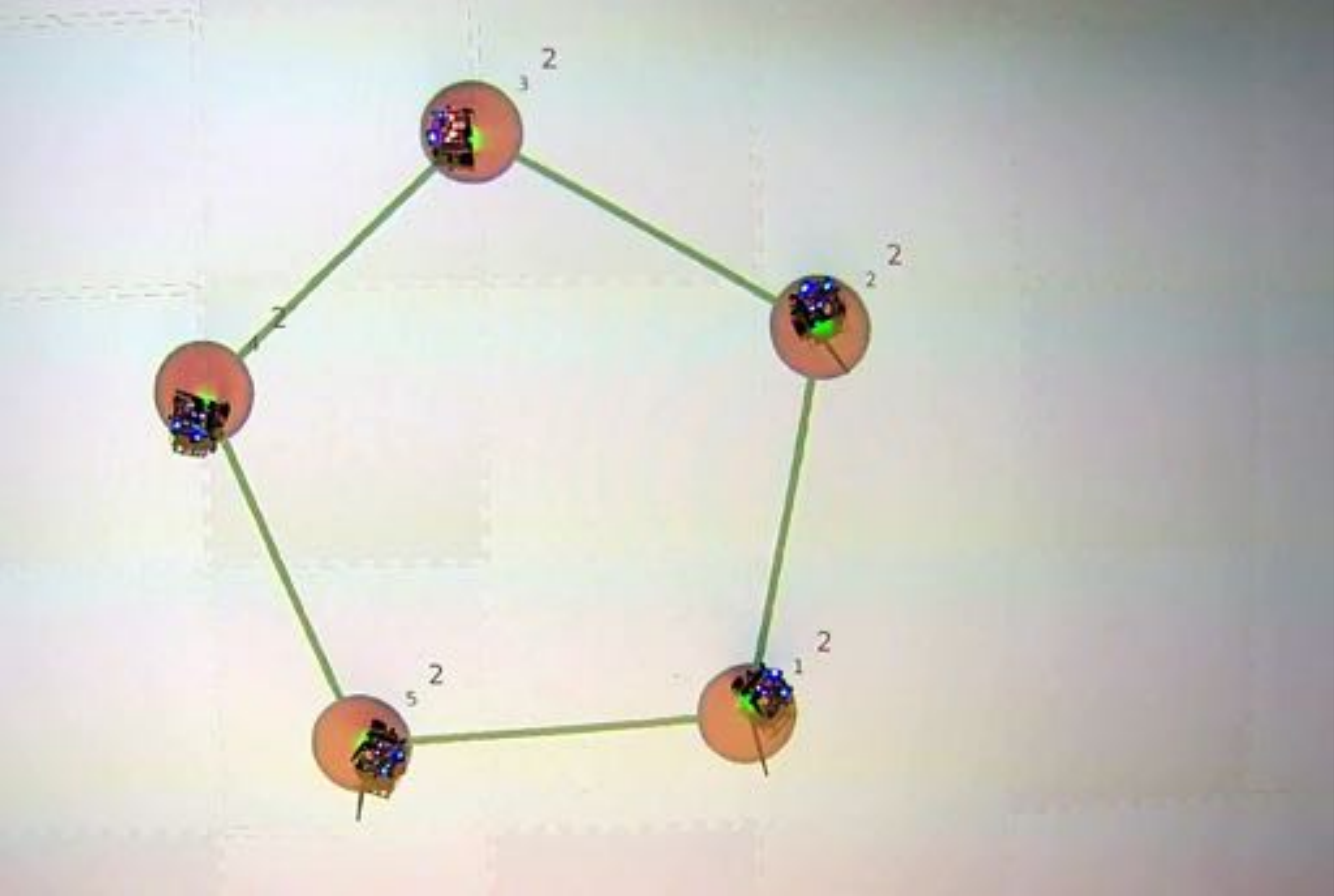}}~
\subcaptionbox{\label{fig2:b}}{\includegraphics[width=0.66\columnwidth, height=0.39\columnwidth]{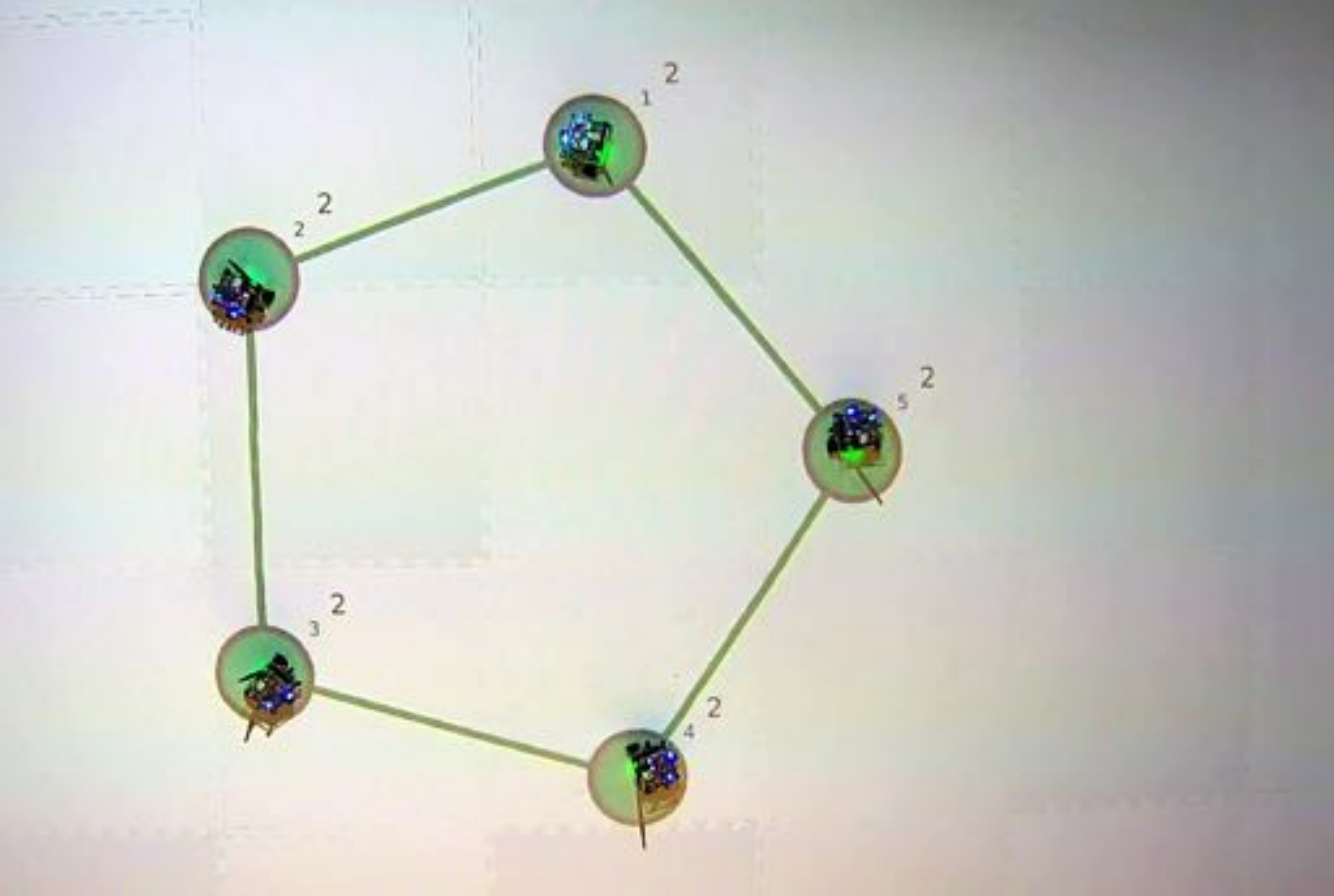}}~
\subcaptionbox{\label{fig2:c}}{\includegraphics[width=0.66\columnwidth, height=0.39\columnwidth]{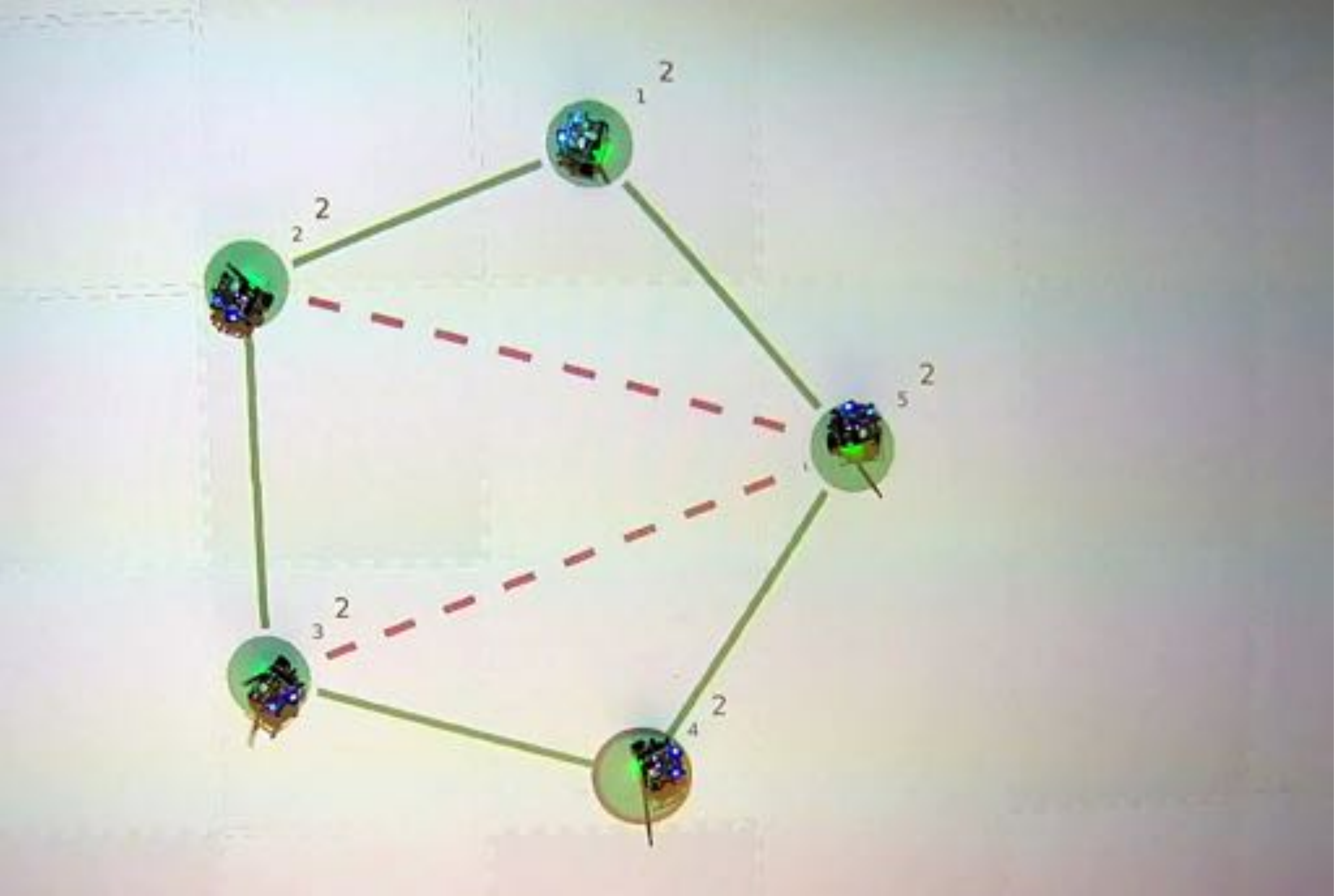}}
} 
\vspace{0.25cm}
\centerline{ 
\subcaptionbox{\label{fig2:d}}{\includegraphics[width=0.66\columnwidth, height=0.39\columnwidth]{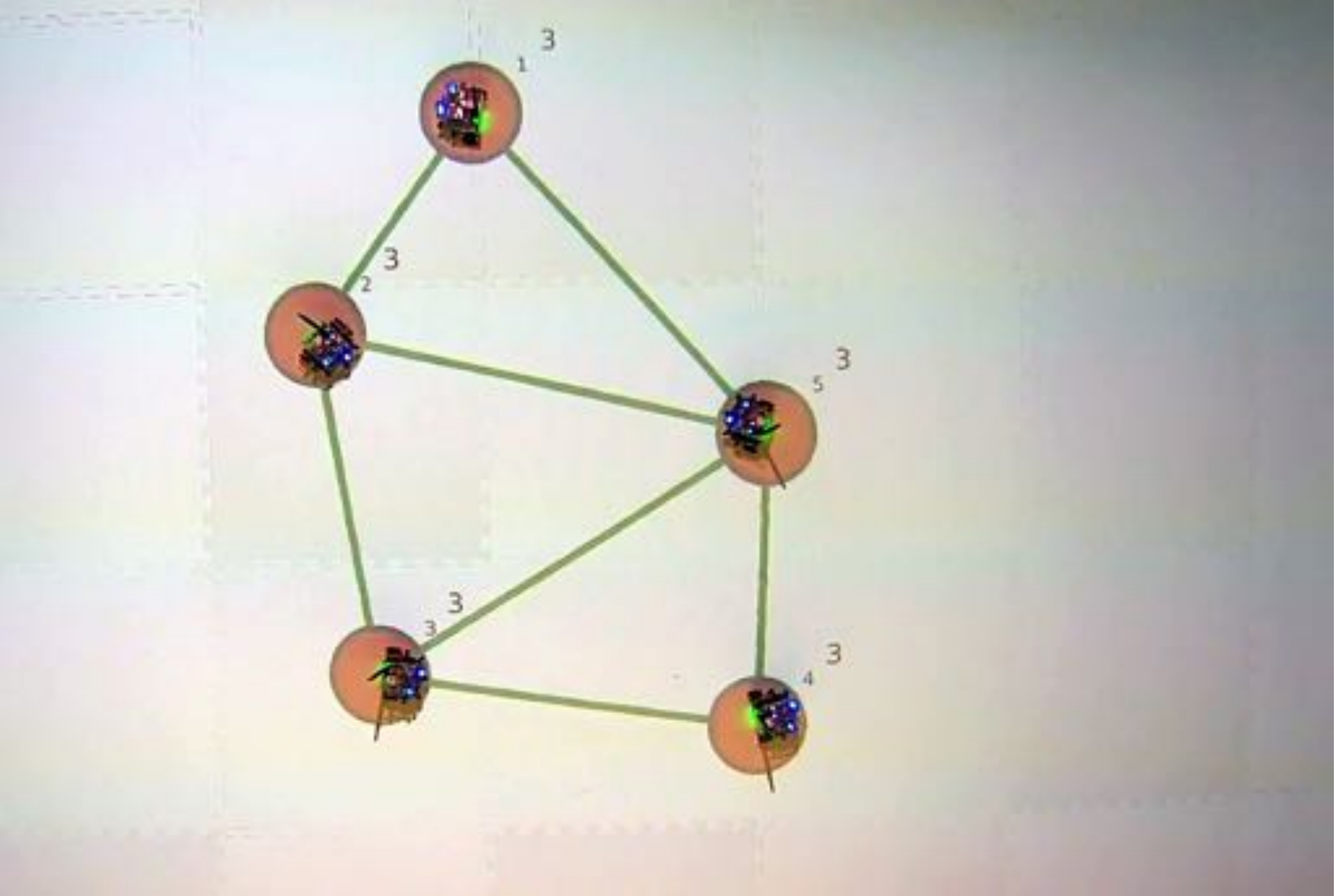}}~
\subcaptionbox{\label{fig2:e}}{\includegraphics[width=0.66\columnwidth, height=0.39\columnwidth]{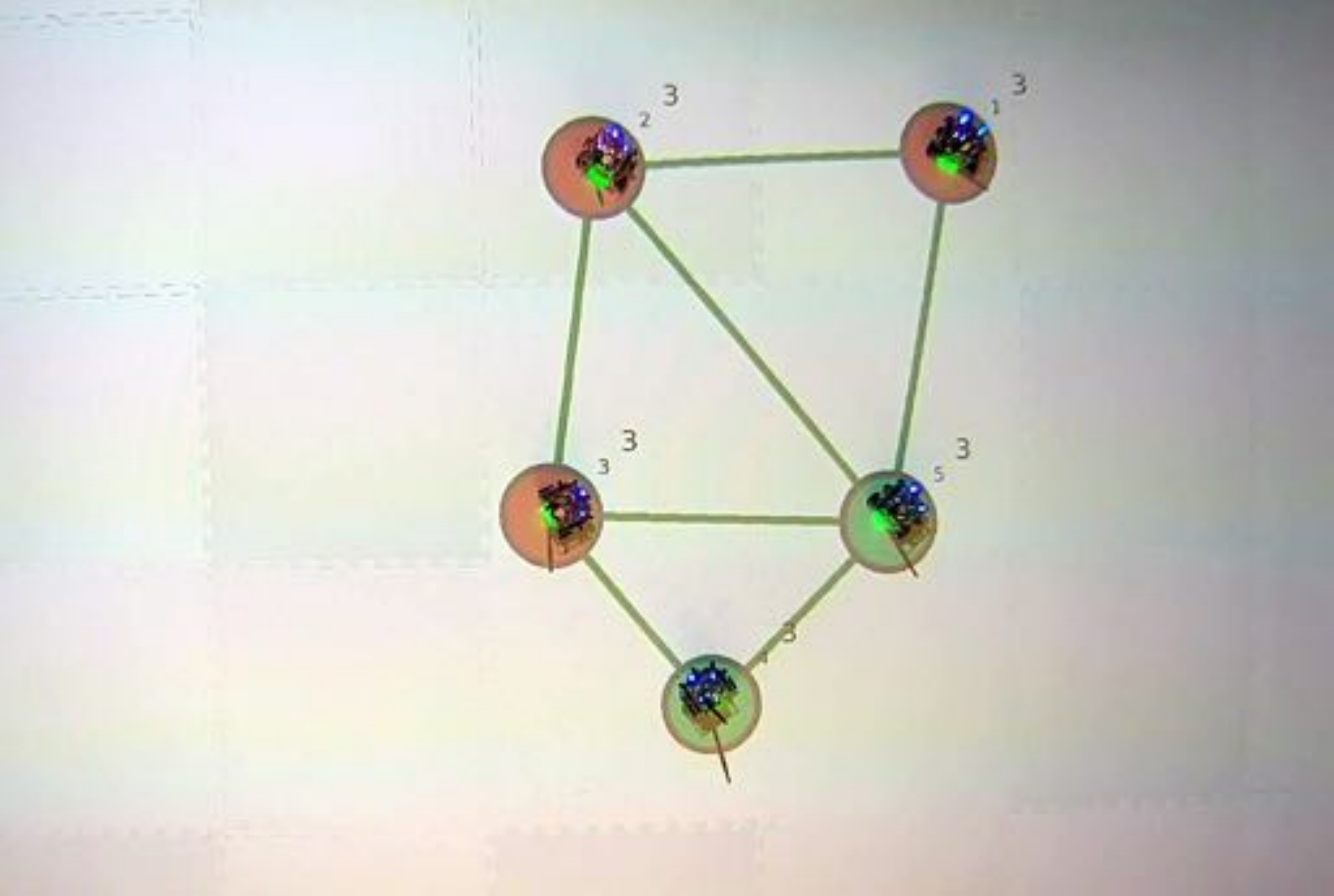}}~
\subcaptionbox{\label{fig2:f}}{\includegraphics[width=0.66\columnwidth, height=0.39\columnwidth]{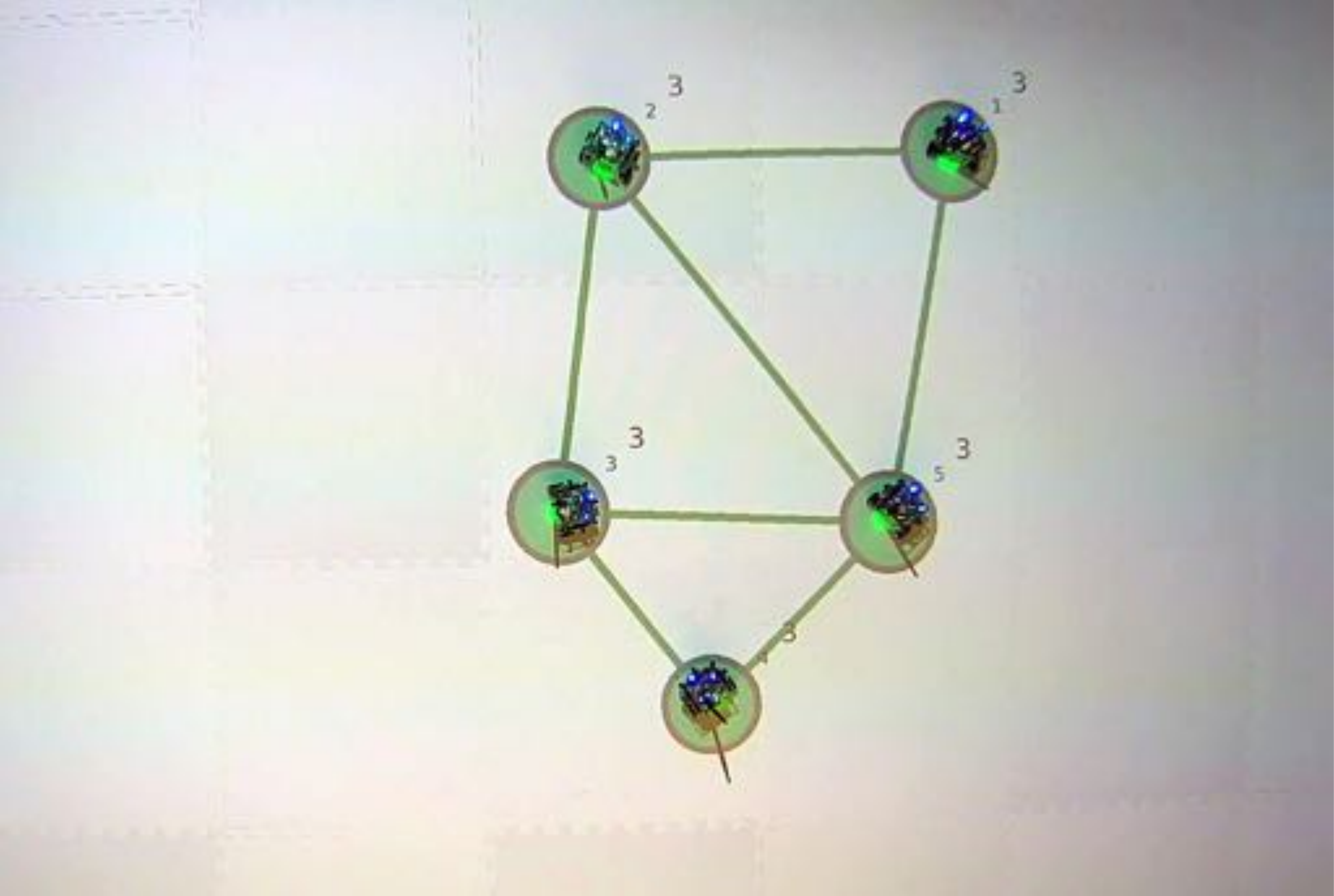}}
}
\caption{Overhead screen-shots from experiments on the Robotarium. Five robots execute two behaviors in sequence (cyclic-pursuit and formation). In figure, green patches represent robots that have completed their task, black rings represent robots that have all neighbors needed for the following task, and green lines represent edges that are available in the current communication graph. From (a) to (b) robots complete the first behavior. During second behavior, additional edges $(2,5)$ and $(3,5)$ are required (red dashed line represent missing edges). From (c) to (d), robots $2,3,5$ reduce their distance below the communication threshold. After the new graph is complete (d), robots initiate the second behavior (e) and complete it (f).
\label{fig:5RobotsExp}}
\end{figure*}

\mymod{We implemented the distributed sequencing framework on the Robotarium~\cite{pickem2017robotarium}, on a team of $5$ differential drive robots. For this example, controllers are designed assuming a single integrator model, i.e. $\bar{f}(x_i)=[0,0]^T$ and $\bar{g}(x_i)=I_2$. In this example, robots execute a transition between two behaviors, where $\mathcal{B}_1$ is a cyclic-pursuit behavior and $\mathcal{B}_2$ is a formation assembly with leader. Cyclic-pursuit behavior is obtained through the following controller:
\begin{equation*}
	\hat{u}_i = \sum_{j \in \mathcal{N}_1^i} R(\phi)\,(x_j - x_i) \quad \forall \, i=1,\dots,5,
\end{equation*}
where $R(\phi)\in SO(2)$ is the rotation matrix of angle $\phi$, which is related to the desired cycle radius. Importantly, for this behavior to work, the communication graph $\mathcal{G}_1$ must be a cycle graph. Considering robot $1$ as leader, the formation control behavior can be achieved with
\begin{align*}
\hat{u}_1 &= \sum_{j \in \mathcal{N}_2^i} \left( (\|x_i-x_j\|^2-\theta_{ij}^2)(x_i - x_i) \right) + \gamma_g(x_g-x_i) \\
	\hat{u}_i &= \sum_{j \in \mathcal{N}_2^i} (\|x_i-x_j\|^2-\theta_{ij}^2)(x_j - x_i) \quad i = 2,\dots 5
\end{align*}
where $\theta_{ij} \in \mathbb{R}_+$ is the desired inter-robot distance, $x_g \in \mathcal{D}$ is the leader's goal, and $\gamma_g \in \mathbb{R}_+$ the corresponding proportional gain. In the case of formation control, it is known that the Euclidean embedding of $\mathcal{G}_2$ must be a rigid framework (see for instance~\cite{mesbahi2010graph} and references therein). With reference to Fig.~\ref{fig:5RobotsExp}, robots initially execute $\mathcal{B}_1$ for a certain amount of time (a). Once completed (b) (green patches represent robots that have completed their current behavior), robots start assembling $\mathcal{G}_2$ (c), after which $\mathcal{B}_2$ is executed until $\| \hat{u}_i \|$ is below a pre-defined threshold (d-f).}

\begin{figure}[h]
    \centering
    \includegraphics[width = \columnwidth]{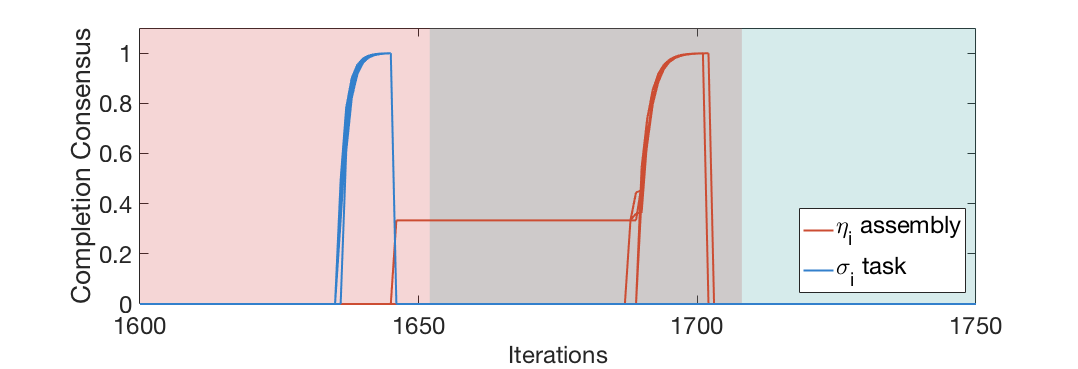}
    \caption{Task and assembly consensus variables $\sigma_i$ and $\eta_i$ for $i=1,\dots,N$ during a transition between two behaviors.}
    \label{fig:transition}
\end{figure}

\mymod{In Fig.~\ref{fig:transition} we can observe the value of the two consensus variables $\sigma_i$ and $\eta_i$ for all robots during the behavior transition. Background colors represent the time intervals during which the two behaviors were executed, while the darker region in the middle corresponds to the assembly of the new graph. We observe the assembly and task variables $\eta_i$ and $\sigma_i$ approaching the value $1$ simultaneously for all robots, thus triggering a synchronized start of the successive phase.}

\begin{figure}[h]
    \centering
    \includegraphics[width = \columnwidth]{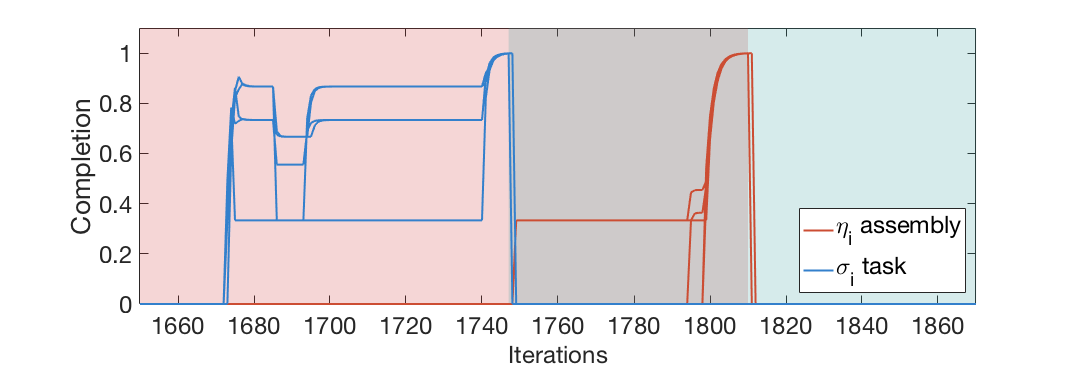}
    \caption{Task and assembly consensus variables $\sigma_i$ and $\eta_i$ for $i=1,\dots,N$ during a transition between two behaviors with communication delays.}
    \label{fig:transition_delay}
\end{figure}

\mymod{The robustness of our technique was tested by simulating uniformly distributed delays between the robots. Results for this case are shown in Fig.~\ref{fig:transition_delay} where we observe that although convergence of $\eta_i$ and $\sigma_i$ is no longer monotonic, robots still reach agreement on when to switch to the successive phase.}

\begin{figure}[h!]
    \centering
    \includegraphics[trim ={2cm 0 0 0}, width=1.05\columnwidth]{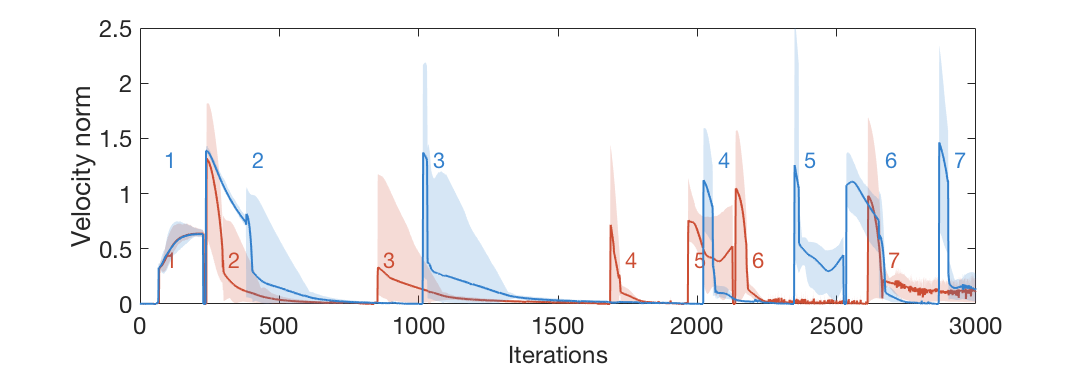}
    \caption{Control input comparison between the minimally invasive sequencing framework proposed in this paper (red) and a sequencing based on rendezvous as {\it gluing} behavior (blue). Solid lines represent the mean of the control input across all robots, while shaded regions represent the interval between minimum and maximum control input.}
    \label{fig:energyComp}
\end{figure}

\mymod{Finally, in order to show the benefits of the minimally invasive approach, we compare it with an alternative technique inspired by~\cite{twu2010graph}, where, upon collective completion of a behavior, robots execute rendezvous until the communication graph required by the successive behavior is assembled. As shown by the simulation results for a sequence of $7$ behaviors (Fig.~\ref{fig:energyComp}), the mean of the input's norm when considering our framework (red solid line) is always lower than the one obtained using the rendezvous as {\it glue} behavior. Importantly, since transitions between behaviors occur faster in the minimally invasive case, the lower control effort cannot be attributed to a more {\it relaxed} choice of controller gains.}


\section{Case Study: Securing a Building} \label{sec:securing}
The objective of this section is to describe the {\it Securing a Building} mission, which will be used as testing scenario for the composition framework. We describe now the main structure and objective of the mission, while we deconstruct it into coordinated behaviors in the next subsection.

\subsection{Mission Overview}
In the Securing a Building mission, a group of robots are deployed in an urban environment to identify an unknown target building and rescue a subject located inside. Based on \cite{FieldManual}, we
decompose this mission into the following 4 phases:

FIND - First, the robots are tasked with identifying the target building by means of surveillance of the perimeters of all the buildings. For efficient exploration, robots can be broken into sub-teams. Each team reports collected information at the base after each building has been investigated. Once the target building has been identified, the robots reunite and prepare for the next phase.

ISOLATE - The robots isolate the target building by patrolling its perimeter. To achieve this, the robots are divided into two subgroups - the {\it security agents} responsible for boundary protection and the {\it maneuvering agents} tasked with entering the building.

RESCUE - During the rescue phase, the security agents keep patrolling around the building. In the meanwhile, the maneuvering agents enter the building, clear the rooms, and seize positions as they maneuver through the building to find the subject to be rescued. Once the subject has been located, the robots transport it to the safe zone.

FOLLOW-THROUGH - As the interior of the building is being cleared, individual robots are left inside as beacons, while the remaining robots from the maneuvering agents leave the building, gather on the outside with the security agents, and report back to the base station.

A number of arguments support the choice of the Securing a Building mission as an ideal scenario for testing multi-robot techniques and algorithms. First, the requirement of spatially diverse functionalities that cannot be provided by single robots naturally requires the use of multi-robot systems. Second, the final goal of the mission, namely rescuing the subjects of interest, reflect the fact that general real-world missions cannot be accomplished with single controllers. Lastly, thanks to its modularity, techniques focusing on specific aspects of the mission can be integrated and tested without influencing the overall structure of the mission (see the Appendix for details).

\subsection{Securing a Building Through Composition of Behaviors}

\begin{figure*}[t]
\centerline{ \includegraphics[width=1.9\columnwidth]{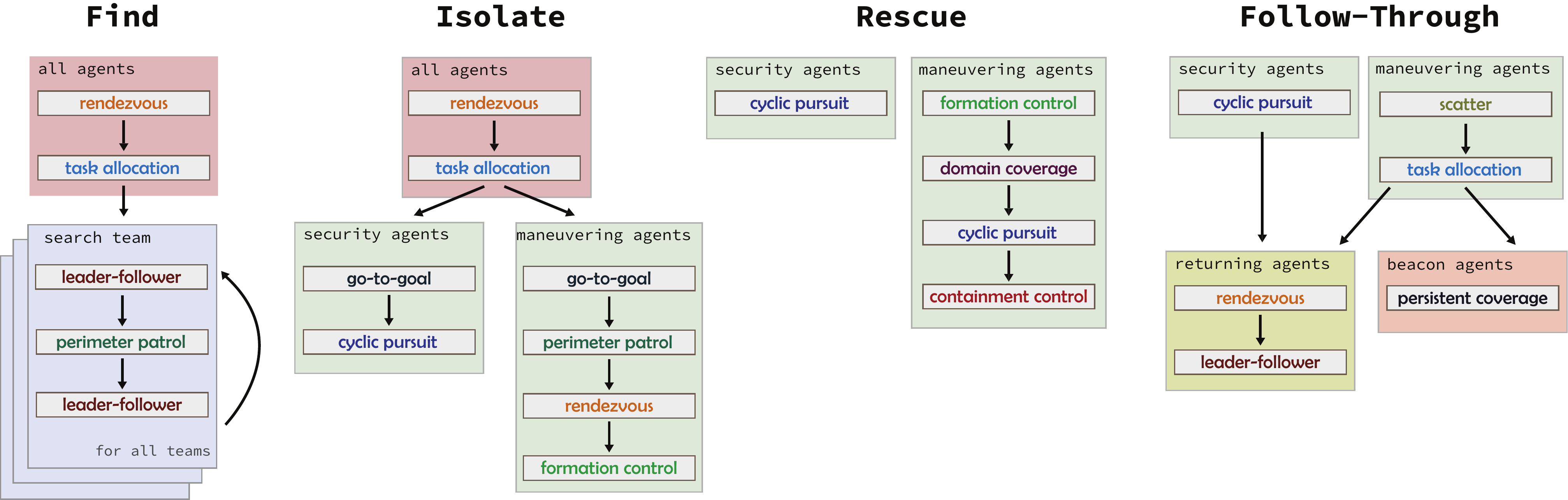}}
\caption{Mission design chart showing how coordinated behaviors are composed together to tackle the Securing a Building mission. The four bold titles are the mission phases and the large boxes below them indicate specific agent roles
and associated behaviors. The arrows in the chart indicate the transitions between different behaviors. We note that he choice of controllers that produces the behaviors in the chart is not unique.\label{fig:missionChart}}
\end{figure*}

We deconstruct the Securing a Building mission through ordered sequences of coordinated behaviors. The process is summarized in Fig.~\ref{fig:missionChart}. We refer to behaviors in terms of their main objectives, acknowledging that different implementations can be used to achieve the same results. We highlight these behaviors in parenthesis.

\paragraph{FIND} Robots initially coordinate with the operator at the base station ({\it rendezvous}). After that, robots are divided into different search teams, each assigned with a list of buildings to investigate ({\it task allocation}). Subsequently, all the teams investigate their own lists of buildings. First team of robots travels to the vicinity of a building ({\it leader-follower}), then start to survey the exterior of the building ({\it perimeter patrol}), and return to the base ({\it leader-follower}). This process repeats until the target building is discovered.

\paragraph{ISOLATE} Robots gather near the base ({\it rendezvous}), then are divided into {\it security} and {\it maneuvering} agents ({\it task allocation}). After traveling from the base to the vicinity of the target building ({\it go-to-goal}), security agents protect the building's perimeter ({\it cyclic pursuit}), until the end of the RESCUE phase. Meanwhile, the maneuvering agents locate the building's entrance, by following its perimeter ({\it perimeter patrol}). Once the entrance has been found, the maneuvering agents gather at the entrance ({\it rendezvous}) and create a formation ({\it formation control}) before entering.

\paragraph{RESCUE} The maneuvering agents enter the building in formation ({\it formation control}) and cover the interior area ({\it area coverage}). Once the location of the subject to rescue is identified, the robots form a circular closure around the subject ({\it cyclic pursuit}). Then, the robots transport the subject to the safety zone, while maintaining the circular closure around the subject ({\it containment control}).

\paragraph{FOLLOW-THROUGH} Maneuvering agents spread ({\it scatter}) over the interior of the building. To signify that the area has been cleared, few robots are left inside the building as beacons ({\it persistent coverage}). The rest of the maneuvering agents and the security agents reunite outside the building ({\it rendezvous}). At last, they return to the base ({\it leader-follower}).

\subsection{Results}
\begin{figure*}[t]
\centerline{ 
\subcaptionbox{\label{fig2:a}}{\includegraphics[width=0.66\columnwidth, height=0.39\columnwidth]{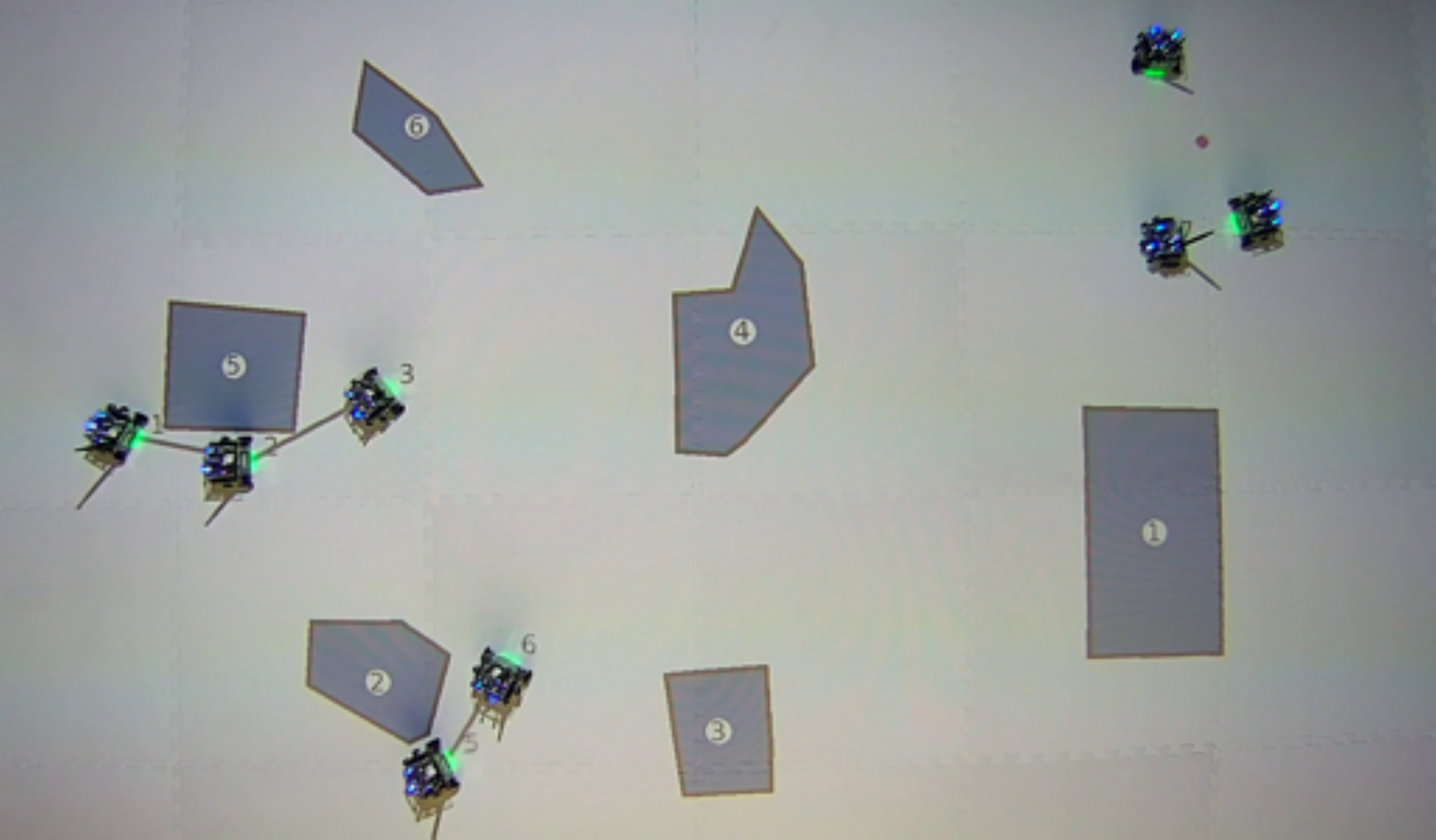}}~
\subcaptionbox{\label{fig2:b}}{\includegraphics[width=0.66\columnwidth, height=0.39\columnwidth]{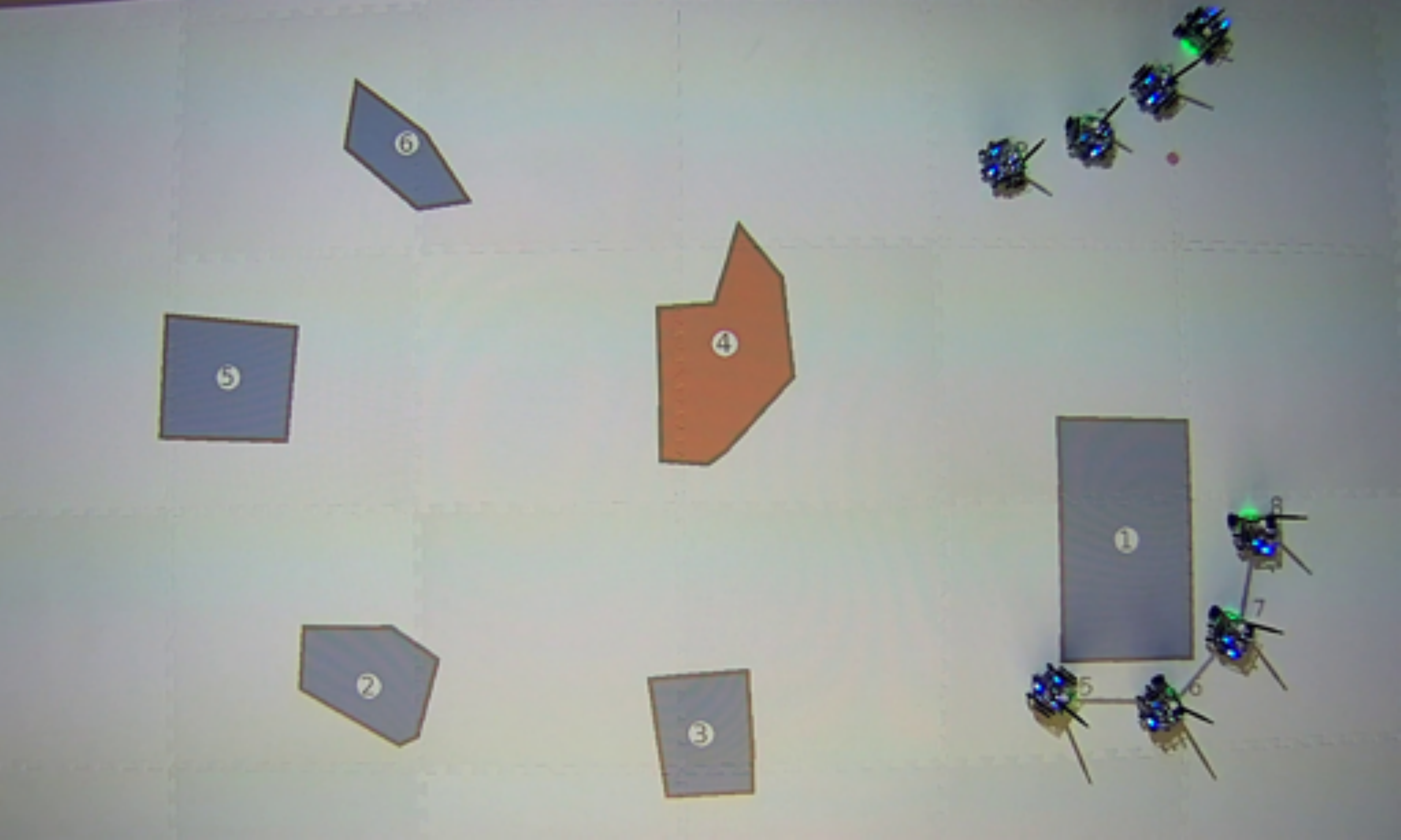}}~
\subcaptionbox{\label{fig2:c}}{\includegraphics[width=0.66\columnwidth, height=0.39\columnwidth]{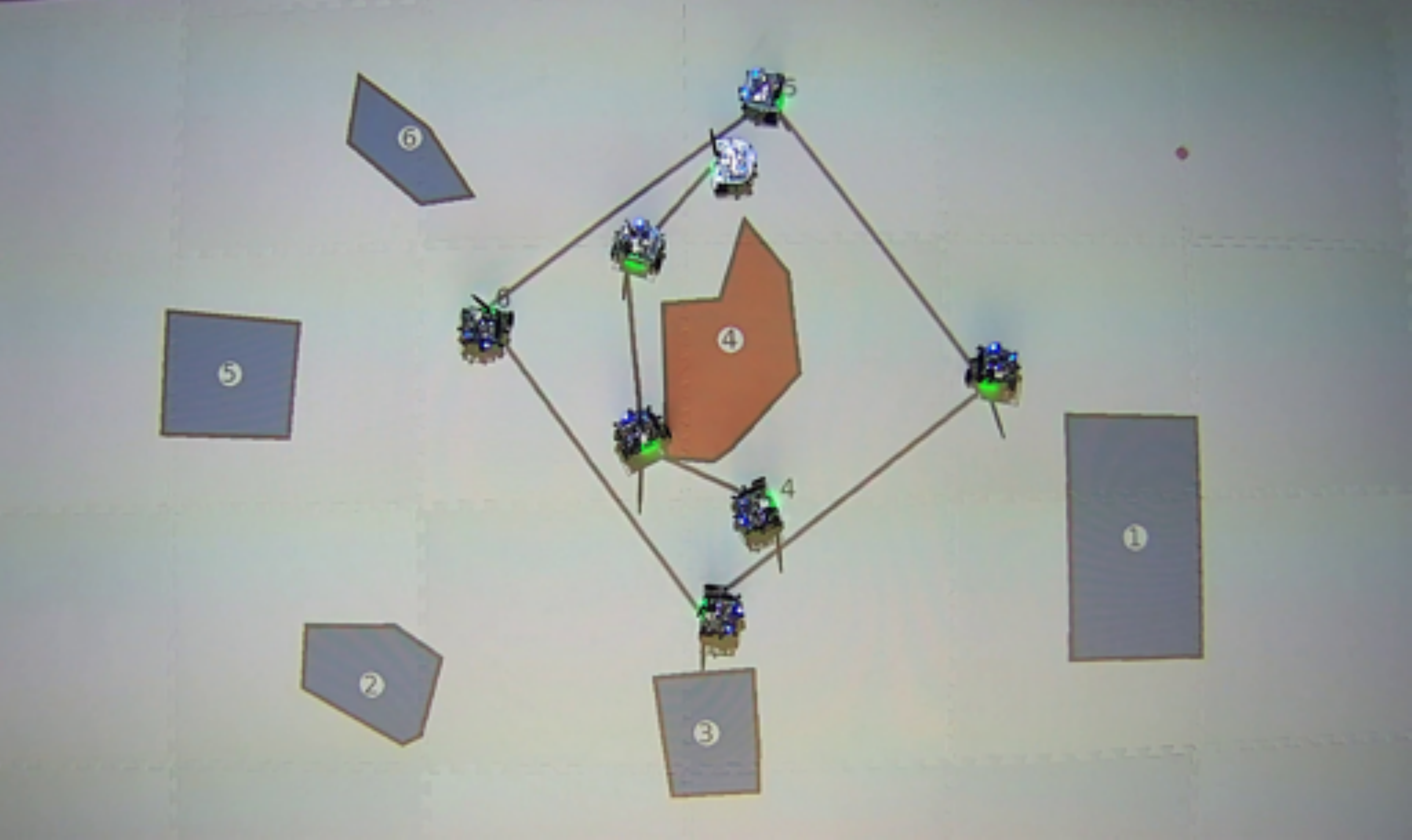}}
} \vspace{0.25cm}
\centerline{ 
\subcaptionbox{\label{fig2:d}}{\includegraphics[width=0.66\columnwidth, height=0.39\columnwidth]{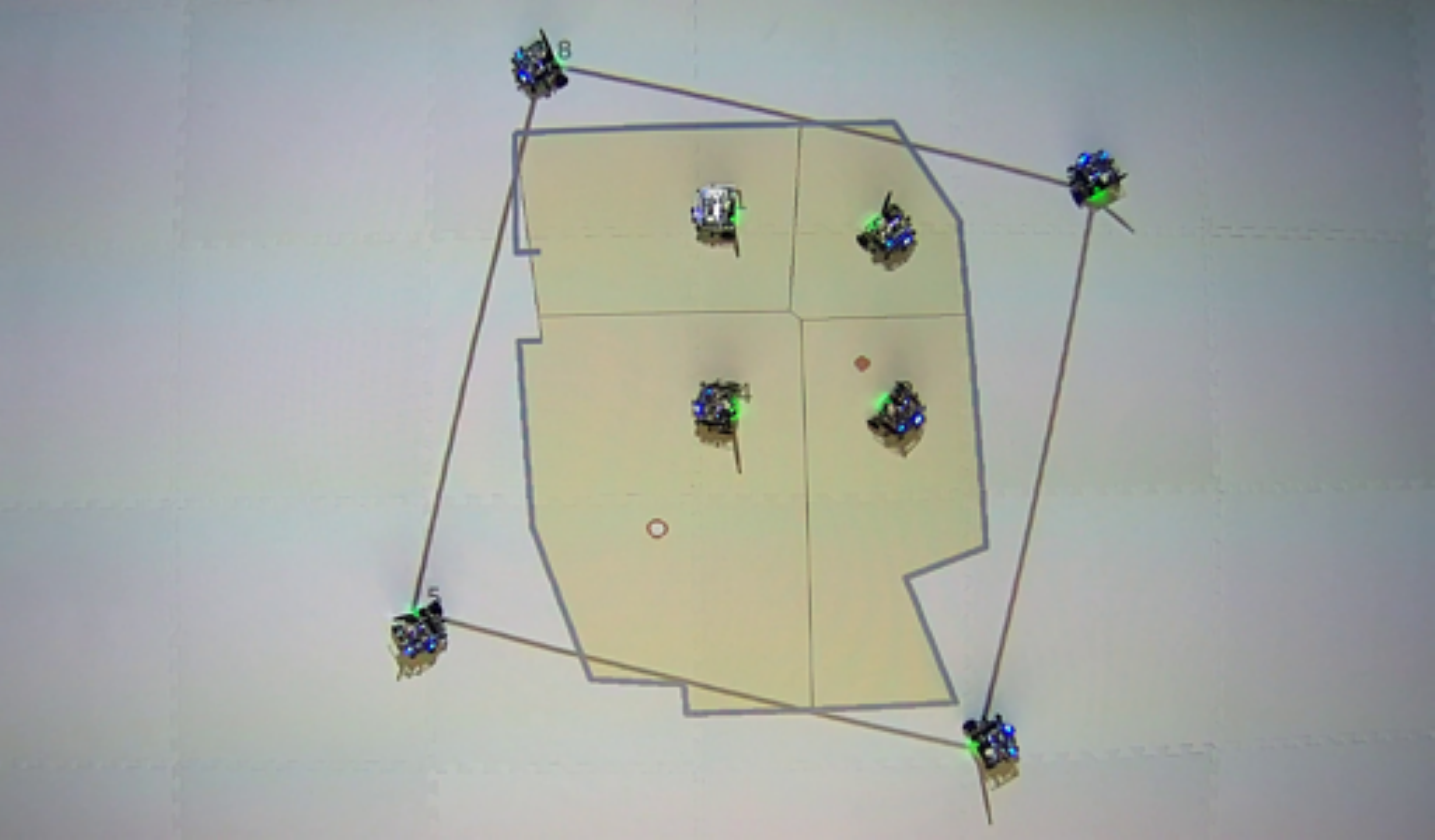}}~
\subcaptionbox{\label{fig2:e}}{\includegraphics[width=0.66\columnwidth, height=0.39\columnwidth]{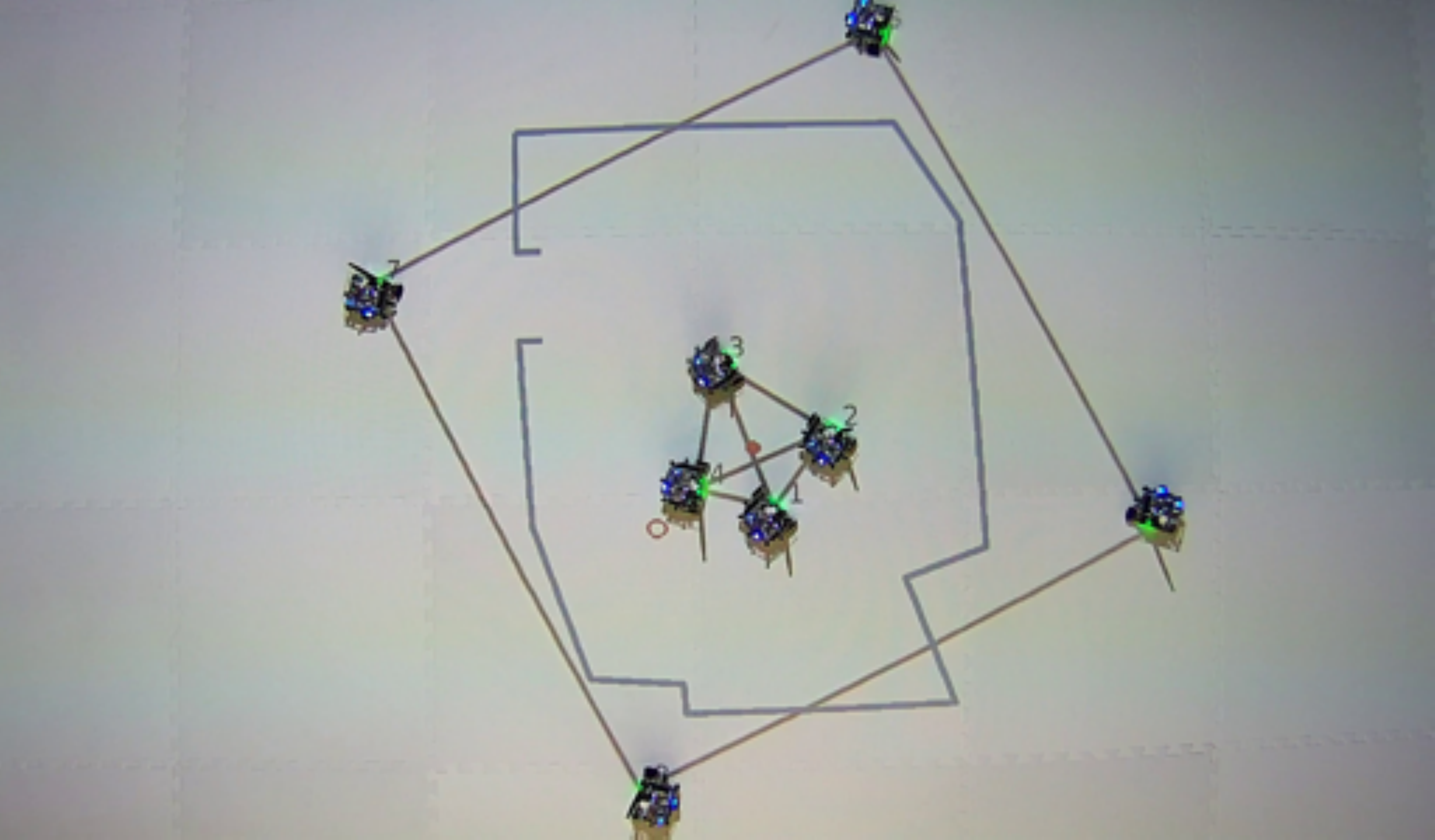}}~
\subcaptionbox{\label{fig2:f}}{\includegraphics[width=0.66\columnwidth, height=0.39\columnwidth]{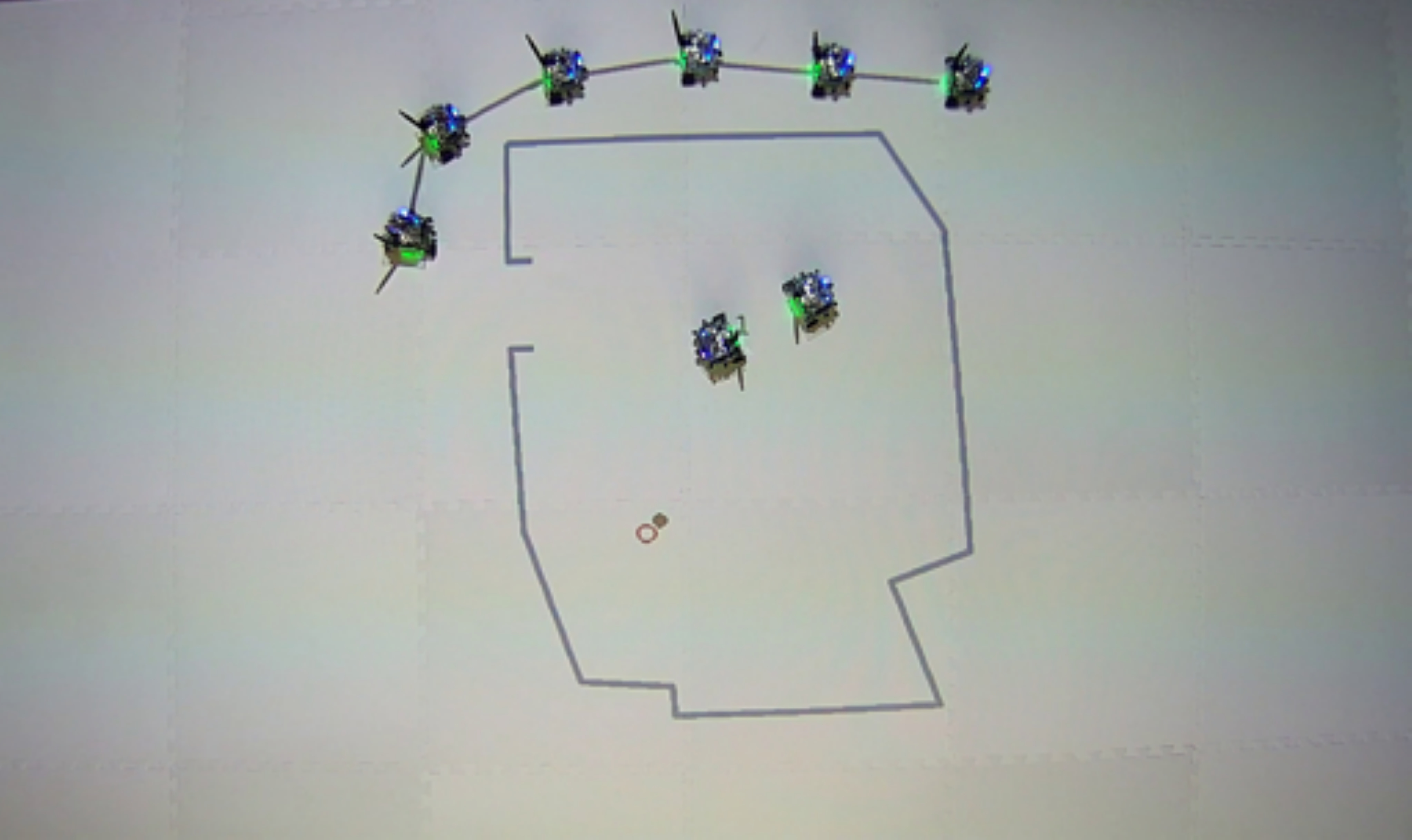}}
}
\caption{Overhead screen-shots from experiments on the Robotarium. A team of eight robots is divided in $\text{\sc team}1:\{1,2,3,4\}$ and $\text{\sc team}2:\{5,6,7,8\}$. Because of the different spatial scales between FIND/ISOLATE phases and RESCUE/FOLLOW-THROUGH phases the mission is executed on two different environments. Each team is assigned with a list of three buildings to inspect sequentially. FIND: (a) perimeter patrol of buildings $2$ and $5$; (b) building $4$ is identified as the target building, while $\text{\sc team}1$ waits for $\text{\sc team}2$ to return to base. ISOLATE: (c) $\text{\sc team}2$ secures perimeter of building, while $\text{\sc team}1$ inspects exterior of building, searching for the entrance. RESCUE: after entering the building, $\text{\sc team}1$ performs domain coverage of the building until target (red dot) is identified (d); after this, (e) robots escort target to safe location (red circle). FOLLOW-THROUGH: finally, two robots are left as beacons inside the building while all remaining robots return to base (f).   
\label{fig:experiment}}

\end{figure*}

We tested the behavior composition framework described in Section~\ref{sec:multAgImp} on the Securing a Building mission, which was executed on the Robotarium~\cite{pickem2017robotarium}. In Fig.~\ref{fig:experiment}, we display selected snapshots of the mission obtained by a camera mounted on the ceiling. In the experiment, $8$ differential-drive robots, indexed $1,\dots,8$ are deployed in a simulated urban environment composed of $6$ buildings, blue polygons indexed $1,\dots,6$. In this experiment, we simulate a maximum sensor range $\Delta = 0.5$m. Because of the different spatial scales between FIND/ISOLATE phases and RESCUE/FOLLOW-THROUGH phases, the entire mission is divided in two parts. In the first part (Fig.~\ref{fig2:a} to Fig.~\ref{fig2:d}) the experiment is performed at a {\it neighborhood}-level scale. The remaining two phases are executed in a zoomed-in environment, which focuses on the one building of interest (Fig.~\ref{fig2:d} to Fig.~\ref{fig2:f}).

During FIND phase (Fig.~\ref{fig2:a} and~\ref{fig2:b}), two groups of robots $\text{\sc team}1:\{1,2,3,4\}$ and $\text{\sc team}2:\{5,6,7,8\}$ investigates preassigned lists of buildings, leaving some agents near the base station (the purple filled dot in the top right corner) if destination building cannot be reached without breaking the connectivity constraints. The red polygon in Fig.~\ref{fig2:b} and~\ref{fig2:c} is the target building after being identified by $\text{\sc team}1$. During the ISOLATE phase (Fig.~\ref{fig2:c}), maneuvering agents look for the entrance, while the security agents secure the outer perimeter. 

During the RESCUE phase (Fig.~\ref{fig2:d} to~\ref{fig2:e}), the agents inside the building, i.e. $\text{\sc team}1$, localize the target (red dot) using Voronoi coverage (Fig.~\ref{fig2:d}) and escort it to the safe area (red circle) as shown in (Fig.~\ref{fig2:e}). Finally, robots $1$ and $2$ are left as beacon inside the building, while remaining robots return to the base (Fig.~\ref{fig2:f}).

\section{Conclusion}
Sequential execution of multi-robot coordinated behaviors can be employed to solve real-world complex missions. However, sequences of behaviors can be executed only if the robots meet all required communication constraints in finite time. In this paper, we described a distributed framework for the sequential composition of coordinated behaviors designed on finite-time convergence control barrier functions. The resulting composition framework is formulated in the form of a quadratic program, which is solved locally by individual robots. \mymod{Although the focus of this paper is on coordinated motion, the application of the proposed framework is relevant to other form of autonomous collaborations where the robots need to satisfy prescribed pair-wise proximity requirements that change over time}. Finally, a large-scale multi-task scenario, denoted ``Securing a Building" mission is proposed as an ideal environment for testing multi-robot techniques.

\appendices
\section{Securing a Building as Benchmark Scenario} \label{sec:appendixA}
Testing the performance of techniques and algorithms for the control of multi-robot systems in real-world scenarios is a challenging task. \mymod{This is particularly true when addressing novel approaches, as the focus on specific aspects of the problem might obscure all-around performance assessments.} To this end, thanks to its modularity, the {\it Securing a Building} mission is an ideal testing framework. In this section, we suggest a number of selected research topics, for which this mission could serve as a testing framework when aiming to evaluate performance of new techniques. \mymod{This appendix is by no mean proposed as a complete list of subjects relevant to multi-robot systems but rather as a discussion to stimulate application of the {\it Securing a Building} as a versatile, real-world testing scenario.}

\paragraph{Team Assembly} Considerable efforts have been devoted to the development of team composition techniques for heterogeneous robots~\cite{prorok2016formalizing},~\cite{koes2005heterogeneous}. Based on the skill set required to solve a particular task, e.g., certain actuation, sensing, locomotion, or communication capabilities, the question is to find a recruitment rule that produces a team capable of delivering the best performance. For instance, in the RESCUE phase, robots capable of opening doors may be required for the {\it maneuvering} agents, while agility and communication capabilities might be preferred during the FIND phase.

\paragraph{Communication} In the context of autonomous networked systems, central roles are played by the flow of information between agents, and the infrastructure required for it~\cite{gupta2016survey}. A number of questions can be posed in relation to the distribution of agents over a domain, given the constraints of communication systems, such as limited range, power requirements, and privacy of the information.

\paragraph{Unknown Environment} The amount of prior knowledge about the environment plays an important role in the definition of both low-level robot controllers and high-level mission plans. The performance of distributed solutions to the localization and mapping problems~\cite{forster2013collaborative} can be tested on the {\it Securing a Building}. Aspect of interest include balancing between exploitation and exploration of the environment applied, for instance, to the building exploration planning.

\paragraph{Resilience} Failure of the mission can be attributed to factors such as damaged components, sensing errors, communication dropouts, delays, control disturbances, reduction of functionalities due to adversarial attacks, etc. A number of different research thrusts focus on the problem of detecting and responding to faults and malicious attacks in multi-agent and cyber-physical systems~\cite{pasqualetti2011consensus,pierpaoli2018fault,fawzi2014secure}.


\bibliographystyle{IEEEtran}
\bibliography{IEEEabrv,biblio.bib}

\begin{thebibliography}{10}
\providecommand{\url}[1]{#1}
\csname url@samestyle\endcsname
\providecommand{\newblock}{\relax}
\providecommand{\bibinfo}[2]{#2}
\providecommand{\BIBentrySTDinterwordspacing}{\spaceskip=0pt\relax}
\providecommand{\BIBentryALTinterwordstretchfactor}{4}
\providecommand{\BIBentryALTinterwordspacing}{\spaceskip=\fontdimen2\font plus
\BIBentryALTinterwordstretchfactor\fontdimen3\font minus
  \fontdimen4\font\relax}
\providecommand{\BIBforeignlanguage}[2]{{%
\expandafter\ifx\csname l@#1\endcsname\relax
\typeout{** WARNING: IEEEtran.bst: No hyphenation pattern has been}%
\typeout{** loaded for the language `#1'. Using the pattern for}%
\typeout{** the default language instead.}%
\else
\language=\csname l@#1\endcsname
\fi
#2}}
\providecommand{\BIBdecl}{\relax}
\BIBdecl

\bibitem{ackerman2014flying}
E.~Ackerman, ``Flying lampshadebots come alive in cirque du soleil
  performance,'' \emph{iEEE Spectrum}, 2014.

\bibitem{du2018fast}
X.~Du, C.~E. Luis, M.~Vukosavljev, and A.~P. Schoellig, ``Fast and in sync:
  Periodic swarm patterns for quadrotors,'' \emph{arXiv preprint
  arXiv:1810.03572}, 2018.

\bibitem{santos2018coverage}
M.~Santos, Y.~Diaz-Mercado, and M.~Egerstedt, ``Coverage control for multirobot
  teams with heterogeneous sensing capabilities,'' \emph{IEEE Robotics and
  Automation Letters}, vol.~3, no.~2, pp. 919--925, 2018.

\bibitem{shishika2018local}
D.~Shishika and V.~Kumar, ``Local-game decomposition for multiplayer
  perimeter-defense problem,'' in \emph{2018 IEEE Conference on Decision and
  Control (CDC)}.\hskip 1em plus 0.5em minus 0.4em\relax IEEE, 2018, pp.
  2093--2100.

\bibitem{han2018hybrid}
H.~Han and R.~G. Sanfelice, ``A hybrid control algorithm for object grasping
  using multiple agents,'' in \emph{2018 IEEE Conference on Control Technology
  and Applications (CCTA)}.\hskip 1em plus 0.5em minus 0.4em\relax IEEE, 2018,
  pp. 652--657.

\bibitem{suarez2011survey}
J.~Suarez and R.~Murphy, ``A survey of animal foraging for directed, persistent
  search by rescue robotics,'' in \emph{2011 IEEE International Symposium on
  Safety, Security, and Rescue Robotics}.\hskip 1em plus 0.5em minus
  0.4em\relax IEEE, 2011, pp. 314--320.

\bibitem{zelazo2018graph}
D.~Zelazo, M.~Mesbahi, and M.-A. Belabbas, ``Graph theory in systems and
  controls,'' in \emph{2018 IEEE Conference on Decision and Control
  (CDC)}.\hskip 1em plus 0.5em minus 0.4em\relax IEEE, 2018, pp. 6168--6179.

\bibitem{cortes2017coordinated}
J.~Cort{\'e}s and M.~Egerstedt, ``Coordinated control of multi-robot systems: A
  survey,'' \emph{SICE Journal of Control, Measurement, and System
  Integration}, vol.~10, no.~6, pp. 495--503, 2017.

\bibitem{lin2003multi}
J.~Lin, A.~S. Morse, and B.~D. Anderson, ``The multi-agent rendezvous
  problem,'' in \emph{42nd IEEE International Conference on Decision and
  Control (IEEE Cat. No. 03CH37475)}, vol.~2.\hskip 1em plus 0.5em minus
  0.4em\relax IEEE, 2003, pp. 1508--1513.

\bibitem{ren2005coordination}
W.~Ren, R.~W. Beard, and T.~W. McLain, ``Coordination variables and consensus
  building in multiple vehicle systems,'' in \emph{Cooperative control}.\hskip
  1em plus 0.5em minus 0.4em\relax Springer, 2005, pp. 171--188.

\bibitem{ramirez2009cyclic}
J.~L. Ramirez, M.~Pavone, and E.~Frazzoli, ``Cyclic pursuit for spacecraft
  formation control,'' in \emph{Proceedings of the American Control
  Conference}, 2009, pp. 4811--4817.

\bibitem{lawton2003decentralized}
J.~R. Lawton, R.~W. Beard, and B.~J. Young, ``A decentralized approach to
  formation maneuvers,'' \emph{IEEE transactions on robotics and automation},
  vol.~19, no.~6, pp. 933--941, 2003.

\bibitem{buckley2017infinitesimally}
I.~Buckley and M.~Egerstedt, ``Infinitesimally shape-similar motions using
  relative angle measurements,'' in \emph{2017 IEEE/RSJ International
  Conference on Intelligent Robots and Systems (IROS)}.\hskip 1em plus 0.5em
  minus 0.4em\relax IEEE, 2017, pp. 1077--1082.

\bibitem{cortes2004coverage}
J.~Cortes, S.~Martinez, T.~Karatas, and F.~Bullo, ``Coverage control for mobile
  sensing networks,'' \emph{IEEE Transactions on robotics and Automation},
  vol.~20, no.~2, pp. 243--255, 2004.

\bibitem{mesbahi2010graph}
M.~Mesbahi and M.~Egerstedt, \emph{Graph theoretic methods in multiagent
  networks}.\hskip 1em plus 0.5em minus 0.4em\relax Princeton University Press,
  2010, vol.~33.

\bibitem{tanner2007flocking}
H.~G. Tanner, A.~Jadbabaie, and G.~J. Pappas, ``Flocking in fixed and switching
  networks,'' \emph{IEEE Transactions on Automatic control}, vol.~52, no.~5,
  pp. 863--868, 2007.

\bibitem{nagavalli2017automated}
S.~Nagavalli, N.~Chakraborty, and K.~Sycara, ``Automated sequencing of swarm
  behaviors for supervisory control of robotic swarms,'' in \emph{2017 IEEE
  International Conference on Robotics and Automation (ICRA)}.\hskip 1em plus
  0.5em minus 0.4em\relax IEEE, 2017, pp. 2674--2681.

\bibitem{ramachandran2019resilience}
R.~K. Ramachandran, J.~A. Preiss, and G.~S. Sukhatme, ``Resilience by
  reconfiguration: Exploiting heterogeneity in robot teams,'' \emph{arXiv
  preprint arXiv:1903.04856}, 2019.

\bibitem{culbertson2018decentralized}
P.~Culbertson and M.~Schwager, ``Decentralized adaptive control for
  collaborative manipulation,'' in \emph{2018 IEEE International Conference on
  Robotics and Automation (ICRA)}.\hskip 1em plus 0.5em minus 0.4em\relax IEEE,
  2018, pp. 278--285.

\bibitem{li2018formally}
A.~Li, L.~Wang, P.~Pierpaoli, and M.~Egerstedt, ``Formally correct composition
  of coordinated behaviors using control barrier certificates,'' in \emph{2018
  IEEE/RSJ International Conference on Intelligent Robots and Systems
  (IROS)}.\hskip 1em plus 0.5em minus 0.4em\relax IEEE, 2018, pp. 3723--3729.

\bibitem{cassandras2009introduction}
C.~G. Cassandras and S.~Lafortune, \emph{Introduction to discrete event
  systems}.\hskip 1em plus 0.5em minus 0.4em\relax Springer Science \& Business
  Media, 2009.

\bibitem{arkin1998behavior}
R.~C. Arkin, \emph{Behavior-based Robotics}.\hskip 1em plus 0.5em minus
  0.4em\relax MIT press, 1998.

\bibitem{koutsoukos2000supervisory}
X.~D. Koutsoukos, P.~J. Antsaklis, J.~A. Stiver, and M.~D. Lemmon,
  ``Supervisory control of hybrid systems,'' \emph{Proceedings of the IEEE},
  vol.~88, no.~7, pp. 1026--1049, 2000.

\bibitem{kress2018synthesis}
H.~Kress-Gazit, M.~Lahijanian, and V.~Raman, ``Synthesis for robots: Guarantees
  and feedback for robot behavior,'' \emph{Annual Review of Control, Robotics,
  and Autonomous Systems}, vol.~1, pp. 211--236, 2018.

\bibitem{srinivasan2018control}
M.~Srinivasan, S.~Coogan, and M.~Egerstedt, ``Control of multi-agent systems
  with finite time control barrier certificates and temporal logic,'' in
  \emph{2018 IEEE Conference on Decision and Control (CDC)}.\hskip 1em plus
  0.5em minus 0.4em\relax IEEE, 2018, pp. 1991--1996.

\bibitem{garg2019control}
K.~Garg and D.~Panagou, ``Control-lyapunov and control-barrier functions based
  quadratic program for spatio-temporal specifications,'' \emph{arXiv preprint
  arXiv:1903.06972}, 2019.

\bibitem{meyer2019hierarchical}
P.-J. Meyer and D.~V. Dimarogonas, ``Hierarchical decomposition of ltl
  synthesis problem for nonlinear control systems,'' \emph{IEEE Transactions on
  Automatic Control}, 2019.

\bibitem{chen2018verifiable}
J.~Chen, S.~Moarref, and H.~Kress-Gazit, ``Verifiable control of robotic swarm
  from high-level specifications,'' in \emph{Proceedings of the 17th
  International Conference on Autonomous Agents and MultiAgent Systems}.\hskip
  1em plus 0.5em minus 0.4em\relax International Foundation for Autonomous
  Agents and Multiagent Systems, 2018, pp. 568--576.

\bibitem{belta2007symbolic}
C.~Belta, A.~Bicchi, M.~Egerstedt, E.~Frazzoli, E.~Klavins, and G.~J. Pappas,
  ``Symbolic planning and control of robot motion [grand challenges of
  robotics],'' \emph{IEEE Robotics \& Automation Magazine}, vol.~14, no.~1, pp.
  61--70, 2007.

\bibitem{klavins2000formalism}
E.~Klavins and D.~E. Koditschek, ``A formalism for the composition of
  concurrent robot behaviors,'' in \emph{2000 IEEE International Conference on
  Robotics and Automation (ICRA)}, vol.~4.\hskip 1em plus 0.5em minus
  0.4em\relax IEEE, 2000, pp. 3395--3402.

\bibitem{marino2009behavioral}
A.~Marino, L.~Parker, G.~Antonelli, and F.~Caccavale, ``Behavioral control for
  multi-robot perimeter patrol: A finite state automata approach,'' in
  \emph{2009 IEEE International Conference on Robotics and Automation
  (ICRA)}.\hskip 1em plus 0.5em minus 0.4em\relax IEEE, 2009, pp. 831--836.

\bibitem{vukosavljev2019hierarchically}
M.~Vukosavljev, A.~P. Schoellig, and M.~E. Broucke, ``Hierarchically consistent
  motion primitives for quadrotor coordination,'' \emph{arXiv preprint
  arXiv:1905.00500}, 2019.

\bibitem{ji2007distributed}
M.~Ji and M.~Egerstedt, ``Distributed coordination control of multiagent
  systems while preserving connectedness,'' \emph{IEEE Transactions on
  Robotics}, vol.~23, no.~4, pp. 693--703, 2007.

\bibitem{sabattini2013distributed}
L.~Sabattini, C.~Secchi, N.~Chopra, and A.~Gasparri, ``Distributed control of
  multirobot systems with global connectivity maintenance,'' \emph{IEEE
  Transactions on Robotics}, vol.~29, no.~5, pp. 1326--1332, 2013.

\bibitem{zavlanos2009hybrid}
M.~M. Zavlanos, H.~G. Tanner, A.~Jadbabaie, and G.~J. Pappas, ``Hybrid control
  for connectivity preserving flocking,'' \emph{IEEE Transactions on Automatic
  Control}, vol.~54, no.~12, pp. 2869--2875, 2009.

\bibitem{igarashi2009passivity}
Y.~Igarashi, T.~Hatanaka, M.~Fujita, and M.~W. Spong, ``Passivity-based
  attitude synchronization in $ se (3) $,'' \emph{IEEE Transactions on Control
  Systems Technology}, vol.~17, no.~5, pp. 1119--1134, 2009.

\bibitem{wang2016multi}
L.~Wang, A.~D. Ames, and M.~Egerstedt, ``Multi-objective compositions for
  collision-free connectivity maintenance in teams of mobile robots,'' in
  \emph{2016 IEEE 55th Conference on Decision and Control (CDC)}.\hskip 1em
  plus 0.5em minus 0.4em\relax IEEE, 2016, pp. 2659--2664.

\bibitem{panerati2019robust}
J.~Panerati, M.~Minelli, C.~Ghedini, L.~Meyer, M.~Kaufmann, L.~Sabattini, and
  G.~Beltrame, ``Robust connectivity maintenance for fallible robots,''
  \emph{Autonomous Robots}, vol.~43, no.~3, pp. 769--787, 2019.

\bibitem{varadharajan2019unbroken}
V.~S. Varadharajan, B.~Adams, and G.~Beltrame, ``The unbroken telephone game:
  Keeping swarms connected,'' in \emph{Proceedings of the 18th International
  Conference on Autonomous Agents and MultiAgent Systems}.\hskip 1em plus 0.5em
  minus 0.4em\relax International Foundation for Autonomous Agents and
  Multiagent Systems, 2019, pp. 2241--2243.

\bibitem{twu2010graph}
P.~Twu, P.~Martin, and M.~Egerstedt, ``Graph process specifications for hybrid
  networked systems,'' \emph{IFAC Proceedings Volumes}, vol.~43, no.~12, pp.
  65--70, 2010.

\bibitem{bhat2000finite}
S.~P. Bhat and D.~S. Bernstein, ``Finite-time stability of continuous
  autonomous systems,'' \emph{SIAM Journal on Control and Optimization},
  vol.~38, no.~3, pp. 751--766, 2000.

\bibitem{xu2015robustness}
X.~Xu, P.~Tabuada, J.~W. Grizzle, and A.~D. Ames, ``Robustness of control
  barrier functions for safety critical control,'' \emph{IFAC-PapersOnLine},
  vol.~48, no.~27, pp. 54--61, 2015.

\bibitem{ames2014control}
A.~D. Ames, J.~W. Grizzle, and P.~Tabuada, ``Control barrier function based
  quadratic programs with application to adaptive cruise control,'' in
  \emph{53rd IEEE Conference on Decision and Control}.\hskip 1em plus 0.5em
  minus 0.4em\relax IEEE, 2014, pp. 6271--6278.

\bibitem{ames2019control}
A.~D. Ames, S.~Coogan, M.~Egerstedt, G.~Notomista, K.~Sreenath, and P.~Tabuada,
  ``Control barrier functions: Theory and applications,'' \emph{2019 European
  Control Conference (ECC)}, pp. 3420--3431, 2019.

\bibitem{squires2019composition}
E.~Squires, P.~Pierpaoli, R.~Konda, S.~Coogan, and M.~Egerstedt, ``Composition
  of safety constraints with applications to decentralized fixed-wing collision
  avoidance,'' \emph{arXiv preprint arXiv:1906.03771}, 2019.

\bibitem{williams2015observability}
R.~K. Williams and G.~S. Sukhatme, ``Observability in topology-constrained
  multi-robot target tracking,'' in \emph{2015 IEEE International Conference on
  Robotics and Automation (ICRA)}.\hskip 1em plus 0.5em minus 0.4em\relax IEEE,
  2015, pp. 1795--1801.

\bibitem{aspnes2006theory}
J.~Aspnes, T.~Eren, D.~K. Goldenberg, A.~S. Morse, W.~Whiteley, Y.~R. Yang,
  B.~D. Anderson, and P.~N. Belhumeur, ``A theory of network localization,''
  \emph{IEEE Transactions on Mobile Computing}, vol.~5, no.~12, pp. 1663--1678,
  2006.

\bibitem{wagenpfeil2009distributed}
J.~Wagenpfeil, A.~Trachte, T.~Hatanaka, M.~Fujita, and O.~Sawodny,
  ``Distributed decision making for task switching via a consensus-like
  algorithm,'' in \emph{2009 American Control Conference}.\hskip 1em plus 0.5em
  minus 0.4em\relax IEEE, 2009, pp. 5761--5766.

\bibitem{pickem2017robotarium}
D.~Pickem, P.~Glotfelter, L.~Wang, M.~Mote, A.~Ames, E.~Feron, and
  M.~Egerstedt, ``The robotarium: A remotely accessible swarm robotics research
  testbed,'' in \emph{2017 IEEE International Conference on Robotics and
  Automation (ICRA)}.\hskip 1em plus 0.5em minus 0.4em\relax IEEE, 2017, pp.
  1699--1706.

\bibitem{FieldManual}
``Military operations in urbanized terrain,'' \emph{US Army Field Manual
  90-10}, 1975.

\bibitem{prorok2016formalizing}
A.~Prorok, M.~A. Hsieh, and V.~Kumar, ``Formalizing the impact of diversity on
  performance in a heterogeneous swarm of robots,'' in \emph{2016 IEEE
  International Conference on Robotics and Automation (ICRA)}.\hskip 1em plus
  0.5em minus 0.4em\relax IEEE, 2016, pp. 5364--5371.

\bibitem{koes2005heterogeneous}
M.~Koes, I.~Nourbakhsh, K.~Sycara, M.~Koes, K.~Sycara, I.~Nourbakhsh, M.~Koes,
  I.~Nourbakhsh, K.~Sycara, S.~D. Ramchurn \emph{et~al.}, ``Heterogeneous
  multirobot coordination with spatial and temporal constraints,'' in
  \emph{AAAI}, vol.~5, 2005, pp. 1292--1297.

\bibitem{gupta2016survey}
L.~Gupta, R.~Jain, and G.~Vaszkun, ``Survey of important issues in uav
  communication networks,'' \emph{IEEE Communications Surveys \& Tutorials},
  vol.~18, no.~2, pp. 1123--1152, 2016.

\bibitem{forster2013collaborative}
C.~Forster, S.~Lynen, L.~Kneip, and D.~Scaramuzza, ``Collaborative monocular
  slam with multiple micro aerial vehicles,'' in \emph{2013 IEEE/RSJ
  International Conference on Intelligent Robots and Systems}.\hskip 1em plus
  0.5em minus 0.4em\relax IEEE, 2013, pp. 3962--3970.

\bibitem{pasqualetti2011consensus}
F.~Pasqualetti, A.~Bicchi, and F.~Bullo, ``Consensus computation in unreliable
  networks: A system theoretic approach,'' \emph{IEEE Transactions on Automatic
  Control}, vol.~57, no.~1, pp. 90--104, 2011.

\bibitem{pierpaoli2018fault}
P.~Pierpaoli, D.~Sauter, and M.~Egerstedt, ``Fault tolerant control for
  networked mobile robots,'' in \emph{2018 IEEE Conference on Control
  Technology and Applications (CCTA)}.\hskip 1em plus 0.5em minus 0.4em\relax
  IEEE, 2018, pp. 374--379.

\bibitem{fawzi2014secure}
H.~Fawzi, P.~Tabuada, and S.~Diggavi, ``Secure estimation and control for
  cyber-physical systems under adversarial attacks,'' \emph{IEEE Transactions
  on Automatic control}, vol.~59, no.~6, pp. 1454--1467, 2014.

\end{thebibliography}

\end{document}